\PassOptionsToPackage{unicode}{hyperref}
\PassOptionsToPackage{hyphens}{url}
\PassOptionsToPackage{dvipsnames,svgnames,x11names}{xcolor}
\documentclass[
  authoryear,
  3p]{elsarticle}

\usepackage{amsmath,amssymb}
\usepackage{iftex}
\ifPDFTeX
  \usepackage[T1]{fontenc}
  \usepackage[utf8]{inputenc}
  \usepackage{textcomp} % provide euro and other symbols
\else % if luatex or xetex
  \usepackage{unicode-math}
  \defaultfontfeatures{Scale=MatchLowercase}
  \defaultfontfeatures[\rmfamily]{Ligatures=TeX,Scale=1}
\fi
\usepackage{lmodern}
\ifPDFTeX\else  
\fi
\IfFileExists{upquote.sty}{\usepackage{upquote}}{}
\IfFileExists{microtype.sty}{% use microtype if available
  \usepackage[]{microtype}
  \UseMicrotypeSet[protrusion]{basicmath} % disable protrusion for tt fonts
}{}
\makeatletter
\@ifundefined{KOMAClassName}{% if non-KOMA class
  \IfFileExists{parskip.sty}{%
    \usepackage{parskip}
  }{% else
    \setlength{\parindent}{0pt}
    \setlength{\parskip}{6pt plus 2pt minus 1pt}}
}{% if KOMA class
  \KOMAoptions{parskip=half}}
\makeatother
\usepackage{xcolor}
\makeatletter
\ifx\paragraph\undefined\else
  \let\oldparagraph\paragraph
  \renewcommand{\paragraph}{
    \@ifstar
      \xxxParagraphStar
      \xxxParagraphNoStar
  }
  \newcommand{\xxxParagraphStar}[1]{\oldparagraph*{#1}\mbox{}}
  \newcommand{\xxxParagraphNoStar}[1]{\oldparagraph{#1}\mbox{}}
\fi
\ifx\subparagraph\undefined\else
  \let\oldsubparagraph\subparagraph
  \renewcommand{\subparagraph}{
    \@ifstar
      \xxxSubParagraphStar
      \xxxSubParagraphNoStar
  }
  \newcommand{\xxxSubParagraphStar}[1]{\oldsubparagraph*{#1}\mbox{}}
  \newcommand{\xxxSubParagraphNoStar}[1]{\oldsubparagraph{#1}\mbox{}}
\fi
\makeatother

\providecommand{\tightlist}{%
  \setlength{\itemsep}{0pt}\setlength{\parskip}{0pt}}\usepackage{longtable,booktabs,array}
\usepackage{calc} % for calculating minipage widths
\usepackage{etoolbox}
\makeatletter
\patchcmd\longtable{\par}{\if@noskipsec\mbox{}\fi\par}{}{}
\makeatother
\IfFileExists{footnotehyper.sty}{\usepackage{footnotehyper}}{\usepackage{footnote}}
\makesavenoteenv{longtable}
\usepackage{graphicx}
\makeatletter
\def\maxwidth{\ifdim\Gin@nat@width>\linewidth\linewidth\else\Gin@nat@width\fi}
\def\maxheight{\ifdim\Gin@nat@height>\textheight\textheight\else\Gin@nat@height\fi}
\makeatother
\setkeys{Gin}{width=\maxwidth,height=\maxheight,keepaspectratio}
\makeatletter
\def\fps@figure{htbp}
\makeatother

\usepackage{booktabs}
\usepackage{longtable}
\usepackage{array}
\usepackage{multirow}
\usepackage{wrapfig}
\usepackage{float}
\usepackage{colortbl}
\usepackage{pdflscape}
\usepackage{tabu}
\usepackage{threeparttable}
\usepackage{threeparttablex}
\usepackage[normalem]{ulem}
\usepackage{makecell}
\usepackage{xcolor}
\usepackage{todonotes,mathtools,bm,amsmath,mathpazo}
\mathtoolsset{showonlyrefs}
\makeatletter
\@ifpackageloaded{caption}{}{\usepackage{caption}}
\AtBeginDocument{%
\ifdefined\contentsname
  \renewcommand*\contentsname{Table of contents}
\else
  \newcommand\contentsname{Table of contents}
\fi
\ifdefined\listfigurename
  \renewcommand*\listfigurename{List of Figures}
\else
  \newcommand\listfigurename{List of Figures}
\fi
\ifdefined\listtablename
  \renewcommand*\listtablename{List of Tables}
\else
  \newcommand\listtablename{List of Tables}
\fi
\ifdefined\figurename
  \renewcommand*\figurename{Figure}
\else
  \newcommand\figurename{Figure}
\fi
\ifdefined\tablename
  \renewcommand*\tablename{Table}
\else
  \newcommand\tablename{Table}
\fi
}
\@ifpackageloaded{float}{}{\usepackage{float}}
\floatstyle{ruled}
\@ifundefined{c@chapter}{\newfloat{codelisting}{h}{lop}}{\newfloat{codelisting}{h}{lop}[chapter]}
\floatname{codelisting}{Listing}

\makeatother
\makeatletter
\@ifpackageloaded{caption}{}{\usepackage{caption}}
\@ifpackageloaded{subcaption}{}{\usepackage{subcaption}}
\makeatother
\journal{International Journal of Production Research}

\ifLuaTeX
  \usepackage{selnolig}  % disable illegal ligatures
\fi
\usepackage[]{natbib}
\usepackage{bookmark}

\IfFileExists{xurl.sty}{\usepackage{xurl}}{} % add URL line breaks if available
\hypersetup{
  pdftitle={A Novel Hybrid Approach to Contraceptive Demand Forecasting: Integrating Point Predictions with Probabilistic Distributions},
  pdfauthor={Harsha Chamara Hewage; Bahman Rostami-Tabar; Aris Syntetos; Federico Liberatore; Glenn R Milano},
  pdfkeywords={family planning supply chain, hybrid
forecasting, forecast distributions, contraceptive demand, forecast
combination},
  colorlinks=true,
  linkcolor={blue},
  filecolor={Maroon},
  citecolor={Blue},
  urlcolor={Blue},
  pdfcreator={LaTeX via pandoc}}

\begin{document}

\begin{frontmatter}
\title{A Novel Hybrid Approach to Contraceptive Demand Forecasting:
Integrating Point Predictions with Probabilistic Distributions}
\author[1]{Harsha Chamara Hewage%
\corref{cor1}%
}
 \ead{HalgamuweHewageHR@cardiff.ac.uk} 
\author[1]{Bahman Rostami-Tabar%
}
 \ead{rostami-tabarb@cardiff.ac.uk} 
\author[2]{Aris Syntetos%
}
 \ead{syntetosa@cardiff.ac.uk} 
\author[3]{Federico Liberatore%
}
 \ead{liberatoref@cardiff.ac.uk} 
\author[4]{Glenn R Milano%
}
 \ead{glenn.milano@gmail.com} 

\affiliation[1]{organization={Cardiff University, Data Lab for Social
Good Group, Cardiff Business School},country={United
Kingdom},countrysep={,},postcode={CF10 3EU},postcodesep={}}
\affiliation[2]{organization={Cardiff University, Cardiff Business
School},country={United Kingdom},countrysep={,},postcode={CF10
3EU},postcodesep={}}
\affiliation[3]{organization={Cardiff University, Cardiff School of
Computer Science \& Informatics},country={United
Kingdom},countrysep={,},postcode={CF24 4AG},postcodesep={}}
\affiliation[4]{organization={United States Agency for International
Development, Bureau for Global Health},country={United
States},countrysep={,},postcode={20523},postcodesep={}}

\cortext[cor1]{Corresponding author}

\begin{abstract}
Accurate demand forecasting is vital for ensuring reliable access to
contraceptive products, supporting key processes like procurement,
inventory, and distribution. However, forecasting contraceptive demand
in developing countries presents challenges, including incomplete data,
poor data quality, and the need to account for multiple geographical and
product factors. Current methods often rely on simple forecasting
techniques, which fail to capture demand uncertainties arising from
these factors, warranting expert involvement. Our study aims to improve
contraceptive demand forecasting by combining probabilistic forecasting
methods with expert knowledge. We developed a hybrid model that combines
point forecasts from domain-specific model with probabilistic
distributions from statistical and machine learning approaches, enabling
human input to fine-tune and enhance the system-generated forecasts.
This approach helps address the uncertainties in demand and is
particularly useful in resource-limited settings. We evaluate different
forecasting methods, including time series, Bayesian, machine learning,
and foundational time series methods alongside our new hybrid approach.
By comparing these methods, we provide insights into their strengths,
weaknesses, and computational requirements. Our research fills a gap in
forecasting contraceptive demand and offers a practical framework that
combines algorithmic and human expertise. Our proposed model can also be
generalized to other humanitarian contexts with similar data patterns.
\end{abstract}

\end{frontmatter}

\section{Introduction}\label{sec-intro}

A fundamental aspect of ensuring reliable access to contraceptive
products lies in accurate demand forecasting, as demand forecasting
forms the foundation of efficient and reliable procurement, sourcing,
storage, allocation, and distribution processes for contraceptive
products \citep{altay2022}. However, the task of producing accurate and
reliable demand forecasts for contraceptives in developing countries
presents numerous challenges \citep{de-arteaga2018}. These challenges
include the unavailability of comprehensive data \citep{lacroix2023},
poor data quality \citep{deboeck2023}, and the necessity to forecast
across multiple geographical and product hierarchies \citep{sedgh2016}.

Despite the complexity inherent in demand forecasting, many forecasts in
practice are often generated using simple methods such as the moving
average of historical consumption data or demographic forecasting
techniques \citep{familyplanninglogisticsmanagement2000}. However, these
methods rarely consider the complexities introduced by users switching
from one contraceptive method to another, driven by factors such as the
introduction of new products, health concerns, or issues with
accessibility and availability \citep{akhlaghi2013}. This inadequacy
contributes to inefficiencies in the family planning supply chain (FPSC)
affecting the availability of contraceptive products \citep{mukasa2017}.

Evidence from the PMA2020 survey\footnote{Performance Monitoring and
  Accountability 2020 survey.} further underscores this issue, revealing
that many health sites and contraceptive outlets in developing countries
often face stockouts of contraceptive methods \citep{ahmed2019}. Such
stockouts limit access to contraceptive products for users when needed,
either by restricting the availability of preferred methods or by
turning away users due to product unavailability \citep{new2017}.
Consequently, these challenges contribute to an increase in unmet
demand\footnote{unmet demand is defined as the percentage of women of
  reproductive age who currently have a need for family planning but are
  either not using any contraceptive methods or whose partners are not
  using them \citep{haakenstad2022}.} for contraceptive products
\citep{Baker2022}.

The unmet demand for contraceptives is a significant concern, as it
leads to an estimated 121 million unintended pregnancies each year,
roughly 331,000 per day \citep{UnitedNations2021}. This situation incurs
substantial costs, both for women and children and for society at large
\citep{sedgh2016}. Over 60\% of unintended pregnancies end in abortion,
whether safe or unsafe, legal or illegal, posing significant risks to
women's lives \citep{bearak2020}. Unfortunately, over 45\% of these
abortions are unsafe and result in maternal deaths \citep{say2014}. This
situation is particularly worse in developing countries, where
approximately 7 million women are hospitalized each year due to unsafe
abortions \citep{singh2016}. Moreover, this also creates a public health
crisis, costing an estimated 2.8 billion USD per year for abortion and
post-abortion care in low- and middle-income countries
\citep{sully2020}. Recognizing its importance, this issue has been
prioritized as essential for achieving the 2023 Sustainable Development
Goals (SDGs).

Despite efforts by governments, foundations, and donors to increase the
uptake of contraceptive products through policy and program
interventions \citep{mukasa2017}, developing countries continue to
experience high unmet demand, particularly due to persistent stockouts
at local health sites and contraceptive outlets \citep{sedgh2014}. A key
reason for this ongoing issue is that these efforts and assessments are
largely focused on the national or global level, which can mask the
ground reality due to local disparities \citep{new2017}.

Recognizing the need for an improved forecasting process at the local
health site level, the United States Agency for International
Development (USAID) launched the \emph{``Intelligent Forecasting
Challenge: Model Future Contraceptive Use''} \citep{USAID2020}. This
competition aimed to source new solutions, test novel ideas, and scale
effective approaches for contraceptive demand forecasting using not only
time series methods but also data driven methods like machine learning
(ML). However, the competition missed a critical element of the
contraceptive demand forecasting process: quantifying and communicating
the uncertainty, as it focused exclusively on point forecasts. On the
other hand, the FPSC in developing countries is often associated with
numerous uncertainties, including complex patterns of demand, variable
lead times, and dependence on donor support \citep{mircetic2022}. These
factors further exacerbate demand uncertainty, necessitating the use of
probabilistic estimations to quantify the uncertainty of future demand.

Discussions with USAID officials highlighted that decision-makers are
particularly interested in the upper bounds of prediction intervals, as
they are keen to mitigate the risk of stockouts, a critical issue in
contraceptive supply chains. As \citet{lacroix2023} has noted, despite
the acknowledged need for probabilistic approaches, point predictions
remain the default due to the lack of standardized methodologies for
incorporating uncertainty into contraceptive demand forecasts.
Probabilistic estimations, by quantifying the uncertainty inherent in
these predictions, thus represent a valuable tool for managing stock
levels and reducing the risk of supply shortages.

To our knowledge, no previous work has focused on probabilistic
forecasting in contraceptive demand estimation within the FPSC at the
local healthcare site level. Thus, this paper first addresses this gap
by investigating a probabilistic forecasting approach for estimating
demand for contraceptive products using data from January 2016 to
December 2019, extracted from the Logistics Management Information
System (LMIS) of Cote d'Ivoire. This is the same dataset that was used
in the competition. Moreover, since the publication of the
\emph{Contraceptive Forecasting Handbook}, which focuses on simple
forecasting methods (see \citet{familyplanninglogisticsmanagement2000}
for more information), there has been no literature evaluating the
applicability and usability of different forecasting methods in
contraceptive demand forecasting. Therefore, in this study, we employ a
range of forecasting methods, including time series, Bayesian, ML and
foundational time series methods\footnote{\textcolor{ForestGreen}{Foundational Time Series Methods: A class of pre-trained machine learning models designed specifically for time series forecasting, leveraging large-scale datasets to generalize across different forecasting tasks. These models require no or minimal tuning and can handle complex temporal patterns efficiently.}},
to produce point forecasts along with probabilistic forecasts for all
products across all healthcare sites.

Additionally, demand planners widely apply judgmental adjustments to
incorporate external factors based on their expertise in the FPSC
setting \citep{altay2022}. Our discussions with USAID professionals
revealed similar insights; they explained that site-level demand
planners often adjust system-generated forecasts or use judgmental
forecasts to eliminate data inaccuracies. These inaccuracies may arise
because the data used to prepare these forecasts may not reflect true
demand due to stockouts or incomplete data collection, or because
planners have additional information, such as product discontinuation
\citep{deboeck2023}. Hence, the human factor is valuable in this
forecasting setup \citep{lacroix2023}.

Given that system-generated forecasts and human forecasts offer distinct
benefits, it is vital to design a hybrid intelligence system that
combines them. In this context, where site-level demand planners produce
point forecasts, we are particularly interested in how to combine point
forecasts with probabilistic forecasts to produce combined probabilistic
forecasts. However, the literature often treats point forecast
combination methods and probabilistic forecast combination methods
separately \citep{wang2023}. To address this gap, we propose a
Constrained Quantile Regression Averaging (CQRA) method\footnote{\textcolor{ForestGreen}{CQRA: A statistical method that combines multiple quantile regression models while enforcing constraints to ensure coherence and interpretability of predictions. It enhances robustness by optimizing weights assigned to different models under quantile-specific constraints, improving accuracy in heterogeneous data settings.}}
to combine point forecasts made by experts with probabilistic forecasts
generated by a system-based forecasting method. We compare the forecast
performance using the Mean Absolute Scaled Error (MASE), a
scale-independent metric that provides robustness and stability for
point forecasts, and Continuous Ranked Probability Scores (CRPS), a
widely used metric in probabilistic forecasting that assesses the
sharpness and calibration of the forecast distribution for probabilistic
forecasts in a cross-validation setup. Finally, we compare our method
results against the submissions from the competition.

Thus, our contributions are as follows:

\begin{enumerate}
\def\labelenumi{\arabic{enumi}.}
\item
  We produce point forecasts along with forecast distributions for
  contraceptive products at the healthcare site level, quantifying
  uncertainties in future contraceptive demand.
\item
  We develop a novel method to combine point forecasts with
  probabilistic forecasts, allowing human experts to incorporate their
  expertise into the forecast and thereby providing a hybrid
  intelligence system.
\item
  We provide a detailed comparison of the performance of time series,
  Bayesian, ML and foundational time series methods, and our proposed
  hybrid method. Additionally, we provide a comparison of the
  computational requirements for each method, offering a holistic view
  of the differences between these forecasting methods.
\item
  We have made the code and data for our proposed method, along with all
  other methods used in this study, publicly accessible to ensure the
  reproducibility. Furthermore, our study adheres to the replication
  principle \citep{boylan2015}, allowing for the method's generalization
  across various sectors with similar data patterns.
\end{enumerate}

The remainder of the paper is structured as follows: Section 2 provides
a brief overview of the literature and discusses its limitations to
position our work. In Section 3, we discuss the data and the
experimental setup. Section 4 presents the results of our analysis. In
Section 5, we summarize our findings, discuss the limitations, and
present ideas for future research directions.

\section{Research background}\label{sec-lit}

Over the past few decades, reducing the unmet need for contraceptives
has been a central focus in the field of FPSC \citep{mukasa2017}. This
issue has been recognized as a critical agenda item in achieving the
2030 Sustainable Development Goals, particularly in expanding access to
contraception to ensure universal access to family planning services
\citep{kantorova2021}. Consequently, accurate and reliable demand
forecasting plays a crucial role in the FPSC, as it supports informed
decision making processes to ensure access to safe and effective
contraceptives, thereby empowering individuals and communities to make
informed reproductive health choices \citep{ahmed2019}.

The majority of literature on forecasting in the FPSC has centered on
estimating family planning indicators at national or global levels to
guide strategic decisions. For instance, \citet{ahmed2019} employed
linear and quadratic logistic regression methods to estimate the modern
contraceptive prevalence rate (mCPR)\footnote{The estimation of the
  percentage of women using a modern contraceptive product
  \citep{ahmed2019}.} in five sub-Saharan African countries, using data
from the PMA2020 survey. Similarly, \citet{new2017} examined trends in
three family planning indicators; 1) mCPR, 2) unmet demand for modern
contraceptives, and 3) demand satisfied by modern contraceptives. Their
study covered the period from 1990 to 2030 for 29 states and union
territories in India. To conduct their analysis, they employed a
Bayesian hierarchical method, integrating statistical time-series
techniques with demographic factors drawn from the Demographic Health
Survey (DHS) \citep{DHSProgram}, Annual Health Survey, and
District-Level Household Survey. \citet{haakenstad2022} used a
spatio-temporal Gaussian process regression method to estimate mCPR,
method mix, and demand satisfied for the global contraceptive prevalence
rate between 1970 and 2019.

These national and global-level studies provide valuable strategic
insights into contraceptive coverage and trends. However, they were not
designed to address operational realities at local levels, such as
variations in demand and supply chain issues at individual service
delivery points. Their focus remains on broader, aggregate-level
decisions that inform national or global policies and strategies
\citep{new2017}. Thus, while these studies may not capture the granular,
site-specific challenges at the operational level, their contributions
are still highly valuable for higher-level planning and resource
allocation.

On the other hand, a fewer have addressed national-level forecasts for
specific contraceptive products. For instance, \citet{akhlaghi2013}
estimated national demand for condoms using demographic-based
forecasting and a consumption-based moving average method, incorporating
expert judgment to refine predictions. \citet{karanja} applied
consumption-based forecasting using Auto Regressive Integrated Moving
Average (ARIMA) and exponential smoothing methods to estimate demand for
contraceptive pills, injectables, implants, and intrauterine
contraceptive devices (IUDs), using data from Kenya's District Health
Information System (DHIS). Moreover, \citet{khan2015} used a
demographic-based forecasting approach integrating expert assumptions
with DHS data to estimate demand for \emph{Sayana Press}, a new
injectable contraceptive, across 12 countries. These studies also
reflect a focus on national or strategic decision-making.

However, at the operational level, where service delivery occurs,
forecasting must account for uncertainties and variations specific to
each location. In contrast to aggregate studies that often incorporate
probabilistic forecasting, which includes prediction intervals to manage
uncertainty \citep{new2017, ahmed2019, haakenstad2022}, operational
studies like \citet{akhlaghi2013} and \citet{karanja} primarily used
point forecasts. These point forecasts provide a single estimate without
explicitly addressing the uncertainty surrounding the forecasted values.
In dynamic settings such as healthcare service delivery, this can limit
their applicability. Operational-level contexts often experience demand
variations due to factors like stockouts, local preferences, and
seasonal changes \citep{mukasa2017}.

Probabilistic forecasts, which include a range of possible outcomes
(such as prediction intervals), are crucial in such contexts because
they acknowledge the uncertainty and help supply chain managers make
more informed decisions \citep{rostami-tabar2024}. For example,
\citet{khan2015} utilized scenario analysis to acknowledge forecast
uncertainty for the injectable contraceptive Sayana Press in 12
countries. However, in the works of \citet{akhlaghi2013} and
\citet{karanja}, only point predictions were generated, with no explicit
consideration of uncertainty in the forecasts. This omission potentially
limits the applicability of their findings in dynamic operational
contexts, as point forecasts, which provide a single value estimate, are
simpler but less adaptable to fluctuating demand conditions.

In practice, however, many still rely on simple methods to produce point
forecasts, despite the limitations these approaches present in capturing
the demand variations for each contraceptive product. \citep{altay2022}.
These methods often fail to address the complexity of demand and attempt
to answer multiple questions using point forecasts \citep{lacroix2023}.
The most commonly employed methods include: 1) extrapolating historical
consumption using basic time series methods, linear trends, averages, or
simple regression methods; 2) estimating consumption based on service
statistics, such as program plans; and 3) utilizing population
demographics to project demand
\citep{familyplanninglogisticsmanagement2000}. In practice, however,
many supply chain managers still rely on point forecasts due to their
simplicity, despite the limitations of these approaches in accounting
for demand fluctuations at the local level \citep{altay2022}. Common
methods include basic time series methods, linear trends, and
demographic-based projections
\citep{familyplanninglogisticsmanagement2000}. While these methods are
easy to implement, they often overlook critical uncertainties and the
complexities of real-world contraceptive demand, such as external shocks
(e.g., global crises like COVID-19), demand shifts due to market
cannibalization, or supply chain disruptions \citep{lacroix2023}.

As a result, these simplistic forecasting approaches can lead to
inefficiencies in ordering and distribution, causing stockouts or
overstocking at healthcare sites, ultimately impairing access to
essential contraceptive services \citep{mukasa2017}. This underscores
the need for more advanced forecasting methods that incorporate both
uncertainty and local variations in demand to optimize supply chain
performance and support better family planning outcomes
\citep{Baker2022}.

\subsection{USAID Intelligent Forecasting
Competition}\label{usaid-intelligent-forecasting-competition}

The USAID Intelligent Forecasting Competition attempted to address the
need for a more reliable contraceptive demand forecasting method by
inviting participants to develop intelligent forecasting methods.
Specifically, participants were tasked with forecasting contraceptive
consumption at the service delivery level of Côte d'Ivoire's public
sector health system over a three-month forecast horizon, using data
provided from the Cote d'Ivoire public health system to forecast the
consumption of contraceptives over three months. This competition
attracted nearly 80 submissions from 40 participants, reflecting a
diverse range of forecasting approaches \citep{USAID2020}.

The winning entry in the USAID Intelligent Forecasting Competition,
developed by Inventec Corporation, used distinct models for each
forecasting horizon. They employed both LightGBM and Long Short-Term
Memory (LSTM) methods, incorporating categorical features, historical
data, and future population projections. The final forecast was derived
from an ensemble of these methods, with weights assigned differently for
each forecasting horizon.

The second-place model employed an ensemble approach, averaging the
predictions from six separate LightGBM models, each designed for a
specific forecasting horizon. This strategy reflects a robust method for
improving forecast accuracy through aggregation. The third-best model
also used LightGBM but incorporated hyperparameter tuning and additional
trend indicators, such as linear and polynomial functions, alongside
various time-series features.

Another notable submission employed hierarchical forecasting with a
bottom-up approach, using ARIMA (AutoRegressive Integrated Moving
Average) as the base forecasting method. This method demonstrates the
utility of hierarchical techniques in complex forecasting scenarios.
Additionally, a model that combined neural networks with a naive
forecasting approach was also among the top contenders. For longer time
series, this model used an ensemble of predictions from different neural
network architectures, including Convolutional Neural Networks (CNN),
Gated Recurrent Units (GRU), and LSTMs, with the final prediction being
the median of these forecasts. For shorter time series, it used a naive
Bayes method. Table~\ref{tbl-literature} presents a summary of the top
10 submissions in the competition\footnote{In this table, the term
  global forecasting refers to a method where a single model is
  developed to handle all time series, while cross-validation refers to
  a technique where the forecasting origin is moved forward by a fixed
  number of steps, producing multiple forecasts at different points in
  time.}.

Although the competition significantly advanced forecasting
methodologies, it primarily concentrated on point forecasts, with
limited attention to the quantification of forecast uncertainty. Since
the competition's objectives did not explicitly include probabilistic
forecasting, criticising these methods for overlooking uncertainty may
be unwarranted. Nevertheless, incorporating uncertainty measures is
crucial for improving forecast reliability, as decision-makers in the
field require dynamic estimates that reflect the uncertainties
associated with contraceptive demand over time \citep{lacroix2023}. The
absence of standard approaches for incorporating uncertainty into
forecasts highlights a significant gap. While the competition
represented a significant step towards the development of advanced
forecasting methodologies, it did not fully address the need for
probabilistic estimations that can better inform decision making by
accounting for forecast variability.

\subsection{Human judgment in contraceptive demand
forecasting}\label{human-judgment-in-contraceptive-demand-forecasting}

In reality, uncertainties in contraceptive demand arise from complex
patterns of product availability, variable lead times, and the
overreliance of donor support \citep{mircetic2022}. These uncertainties
are further complicated by inadequacies in data collection, storage, and
sharing practices. Despite the implementation of Logistics Management
Information Systems (LMIS), field operatives frequently depend on
paper-based forms and Excel spreadsheets for data collection and
operational management \citep{devries2020}. This reliance on outdated
methods often results in noisy, inaccurate, and incomplete data, thereby
complicating the forecasting efforts \citep{besiou2020}. Moreover,
stockout driven consumption data may not reflect actual demand, further
distorting the forecasting process \citep{deboeck2023}.

Forecasting algorithms, while adept at processing high-dimensional data,
may struggle to detect these sudden fluctuations and discontinuities in
contraceptive demand \citep{hong2021}. Thus, this leaves the expert to
use their contextual knowledge to understand the context of the data and
incorporate it with the forecasting process \citep{hong2021}. Previous
literature also suggests that experts can improve the forecast
performance when the external information has not been added to the
algorithm-based forecasting method using their inside knowledge and
expertise\footnote{See \citet{perera2019} for a detailed review on human
  factors in supply chain forecasting.}
\citep{fildes2007, davydenko2013}.

In the context of FPSC in developing countries, current algorithm-based
forecasting methods possess the capability to manage extensive datasets,
including thousands of time series across diverse geographies
simultaneously \citep{hong2021}. However, these algorithms may not fully
account for external factors and contextual nuances that experts are
adept at identifying \citep{devries2020}. This suggests that integrating
human expertise with algorithm-based forecasts could yield more reliable
and accurate predictions. Empirical evidence also supports the notion
that combining human judgment with algorithmic forecasts enhances
forecast performance \citep{fildes2009, petropoulos2018}. Furthermore,
literature shows that such forecast combinations improve forecasting
accuracy compared to using either method in isolation \citep{wang2023}.
This leads to the critical question of how best to integrate human
expertise with algorithm-based forecasting.

\textcolor{ForestGreen}{Various approaches exist for integrating human judgment with algorithmic forecasts, forming hybrid intelligence systems that leverage the strengths of both.}
\citet{brau2023}
\textcolor{ForestGreen}{categorize these methods into five key types:}

\begin{itemize}
\tightlist
\item
  \emph{Judgmental Adjustment} --
  \textcolor{ForestGreen}{Experts modify algorithmic forecasts based on contextual insights.}\\
\item
  \emph{Quantitative Correction} --
  \textcolor{ForestGreen}{Systematic adjustments are applied to human forecasts using statistical techniques.}\\
\item
  \emph{Forecast Combination} --
  \textcolor{ForestGreen}{Separate judgmental and algorithm-based forecasts are merged into a single forecast.}\\
\item
  \emph{Judgment as a Model Input} --
  \textcolor{ForestGreen}{Expert knowledge is incorporated as a predictive variable within the forecasting model.}\\
\item
  \emph{Integrative Judgment Learning} --
  \textcolor{ForestGreen}{Human inputs iteratively refine model predictions through a structured learning process.}\\
\end{itemize}

\textcolor{ForestGreen}{Judgmental adjustments have been widely studied in supply chain contexts, particularly in promotional settings.}
\citet{trapero2013}
\textcolor{ForestGreen}{found that structured human interventions during promotions improved forecasting accuracy. This aligns with findings from}
\citet{fildes2015},
\textcolor{ForestGreen}{who argue that human adjustments enhance reliability when bias is minimized.}

\textcolor{ForestGreen}{Quantitative correction methods offer a structured way to adjust forecasts while maintaining statistical integrity.}
\citet{fildes2009}
\textcolor{ForestGreen}{highlight the importance of systematic corrections to mitigate bias and improve forecast accuracy.}

\textcolor{ForestGreen}{Forecast combination has been explored as a way to merge expert forecasts with algorithmic outputs.}
\citet{goodwin2000}
\textcolor{ForestGreen}{highlights its effectiveness in capturing both statistical trends and domain knowledge, making it a useful technique when expert input is available alongside data-driven predictions.}

\textcolor{ForestGreen}{Another structured approach to integrating expert knowledge into forecasting models is treating human judgment as a direct input to model-building. Instead of adjusting forecasts post hoc, this method incorporates expert-driven insights as predictive variables, allowing the model to factor in domain knowledge systematically.}
\citet{arvan2019}
\textcolor{ForestGreen}{suggest that using expert judgment as a model input can improve forecasting accuracy, particularly in cases where statistical models alone struggle to capture contextual nuances. By embedding domain expertise within the forecasting process, this approach ensures that human insights are systematically integrated, rather than relying on manual interventions.}

\textcolor{ForestGreen}{On the other hand, integrative Judgment Learning}
\citet{baecke2017}, \citet{brau2023}
\textcolor{ForestGreen}{treats human adjustments as predictive variables, allowing models to systematically weigh their impact. Similarly,}
\citet{goodwin1999}
\textcolor{ForestGreen}{suggest that hybrid forecasting models should be designed to correct systematic biases in human judgment while maintaining adaptability.}

\textcolor{ForestGreen}{Some studies explore hybrid forecasting methods that do not fit neatly into a single category. These approaches often involve adaptive weighting mechanisms}
\textcolor{ForestGreen}{optimization-driven model adjustments},
\textcolor{ForestGreen}{or learning-based frameworks that dynamically integrate human and machine-generated forecasts}
\citep{petropoulos2022}.
\textcolor{ForestGreen}{While not explicitly categorized under the five methods, these strategies contribute to the broader understanding of how expert judgment and algorithmic forecasts can be effectively combined.}

In this context, two key considerations arise when selecting an approach
for combining forecasts: 1) healthcare staff at the site level in
developing countries often lack the specialized skills and training
needed to develop and maintain sophisticated forecasting methods
\citep{altay2022}, and 2) based on discussions with USAID officials,
healthcare staff typically produce only point forecasts, often relying
on their judgment, as also highlighted in the literature
\citep{akhlaghi2013}. Moreover, \citet{lacroix2023} found that although
practitioners acknowledge the importance of accounting for uncertainty,
there is no widely adopted standard for doing so. Judgmental point
forecasts are favored because they align with the practical experience
and cognitive abilities of healthcare staff, who may find it difficult
to quantify uncertainty without formal training in probabilistic ms.
Given these constraints, combining human judgmental forecasts with
algorithmic forecasts presents a promising solution. Empirical studies
consistently show that combining forecasts, whether judgmental or
model-based, generally improves accuracy over relying on individual
forecasts alone \citep{ranjan, wang2023}. However, this introduces a
challenge: how to effectively integrate human point forecasts with
probabilistic forecasts from algorithms to produce a unified
probabilistic forecast. Although the literature on forecast combinations
is extensive, little attention has been paid to the integration of point
forecasts with probabilistic ones (see \citet{wang2023} for a
comprehensive review).

\subsection{Literature limitations
summary}\label{literature-limitations-summary}

\textcolor{ForestGreen}{Despite advancements in contraceptive demand forecasting, several key gaps remain unaddressed. First, the use of probabilistic forecasting in FPSC remains limited. While probabilistic methods have been widely explored in other domains, such as economic forecasting}
\citep{gneiting2014, krüger2017},
\textcolor{ForestGreen}{their application in FPSC is largely absent. Most existing studies rely on point forecasts, which do not account for demand uncertainty, making supply chain planning more vulnerable to unexpected fluctuations. Given the inherent unpredictability of contraceptive demand, incorporating probabilistic forecasting is critical for improving stock management and mitigating stockout risks}

\textcolor{ForestGreen}{Second, contraceptive demand forecasting has primarily been conducted at the national or regional level}
\citep{akhlaghi2013, karanja},
\textcolor{ForestGreen}{often overlooking demand variability at individual healthcare sites. However, demand patterns vary significantly across locations}
\citep{karimi2021},
\textcolor{ForestGreen}{and without localized forecasts, procurement decisions may fail to align with actual site-level needs. This highlights the need for forecasting approaches that can adapt to site-specific demand while maintaining consistency with broader supply chain strategies.}

\textcolor{ForestGreen}{Third, while forecast combination techniques have been extensively studied}
\citep{ranjan},
\textcolor{ForestGreen}{their application in FPSC is underexplored. Existing studies typically assess individual forecasting methods without considering how integrating multiple forecasts could improve accuracy, particularly in healthcare supply chains. Additionally, there is limited empirical research comparing different forecasting methods for contraceptive demand estimation, making it difficult to determine the most effective approach in this context.}

\textcolor{ForestGreen}{Finally, although judgmental adjustments are widely used in FPSC}
\citep{fildes2009, trapero2013},
\textcolor{ForestGreen}{there is no structured approach that systematically integrates expert knowledge while ensuring coherence between human forecasts and probabilistic distributions in FPSC settings. In economic forecasting, Entropic Tilting has been used to refine probabilistic distributions based on external point forecasts}
\citep{krüger2017, metaxoglou2016},
\textcolor{ForestGreen}{yet these methods do not explicitly ensure coherence between expert judgment and forecast distributions. Our study addresses this gap by proposing a hybrid forecasting approach that systematically integrates expert forecasts with probabilistic models, ensuring that adjustments remain statistically grounded while incorporating expert insights.}

\begin{landscape}\begingroup\fontsize{9}{11}\selectfont

\begin{longtable}[t]{>{\raggedright\arraybackslash}p{8em}>{\raggedright\arraybackslash}p{18em}>{\raggedright\arraybackslash}p{6em}>{\raggedright\arraybackslash}p{9em}>{\raggedright\arraybackslash}p{6em}>{\raggedright\arraybackslash}p{6em}>{\raggedright\arraybackslash}p{6em}}

\caption{\label{tbl-literature}Summary of the top 10 models in the USAID
intelligent forecasting competition.}

\tabularnewline

\toprule
Reference & Method & Metric & Forecasting strategy & Probabilistic & Global forecasting & Cross validation\\
\midrule
Current study & sNAIVE,  Moving average, Exponential Smoothing State Space, ARIMA, Syntetos-Boylan approximation, Bayesian structural time series, Multiple Linear Regression, LightGBM, xgBoost, Random Forest, TimeGPT, Chronos, Lag Llama, Demographic, Forecast combination models and Proposed hybrid method & MASE, CRPS & Recursive multi-step forecasting & YES & YES & YES\\
Submission 1 & An ensemble model of LightGBM and LSTM using weighted average & MASE & Direct multi-step forecasting & NO & YES & NO\\
Submission 2 & LSTM model & MASE & Direct multi-step forecasting & NO & YES & NO\\
Submission 3 & A simple ensemble of six LightGBM models & MASE & Direct multi-step forecasting & NO & YES & NO\\
Submission 4 & LightGBM model & MASE & Direct multi-step forecasting & NO & YES & NO\\
\addlinespace
Submission 5 & LightGBM model & MASE & Recursive multi-step forecasting & NO & YES & NO\\
Submission 6 & A simple ensemble of six LightGBM models & MASE & Direct multi-step forecasting & NO & YES & NO\\
Submission 7 & A simple ensemble of LightGBM and LSTM & MASE & Recursive multi-step forecasting & NO & YES & NO\\
Submission 8 & A simple ensemble of three LightGBM models & MASE & Direct multi-step forecasting & NO & YES & NO\\
Submission 9 & Hierarchical timeseries model using ARIMA & MASE & Recursive multi-step forecasting & NO & NO & NO\\
\addlinespace
Submission 10 & A simple ensemble of nine LSTM models and Naïve Bayes model & MASE & Recursive multi-step forecasting & NO & YES & NO\\
\bottomrule

\end{longtable}

\endgroup{}
\end{landscape}

\section{Proposed hybrid approach}\label{sec-model}

We propose a CQRA model to generate a combined probabilistic forecast,
utilizing both point and probabilistic forecasts. This approach builds
upon the CQRA model introduced by \citet{wang_combining_2018}, which
focuses on combining multiple probabilistic forecasts to produce a
consolidated forecast distribution. The key concept in our proposed
method is to generate quantiles from a given probabilistic forecast and
adjust each quantile using weights. These weights are determined by
treating the point forecast as the ``new reality'' and formulating a
linear programming (LP) problem that minimizes both the pinball
loss\footnote{\textcolor{ForestGreen}{Pinball Loss: A proper scoring rule used to evaluate quantile forecasts, penalizing deviations based on whether the predicted quantile overestimates or underestimates the observed value. It ensures a well-calibrated probabilistic forecast by emphasizing accuracy across different quantile levels.}}
and the absolute error between point forecast and mean of the weighted
quantile forecast. The pinball loss is a strictly proper scoring rule
used to evaluate quantile forecasts. It measures overall quantile
performance by rewarding sharpness and penalizing miscalibration
\citep{hyndman2021forecasting}.

\textcolor{ForestGreen}{On the other hand, our proposed CQRA model shares similarities with the quantile combination approach of}
\citet{trapero2019}
\textcolor{ForestGreen}{in that both methods determine weights by minimizing the tick-loss function. However, a fundamental difference is that while}
\citet{trapero2019}
\textcolor{ForestGreen}{combine quantile forecasts generated from Kernel Density Estimators (KDE) and GARCH models to optimize safety stock levels, CQRA integrates expert point forecasts with probabilistic forecasts. This allows CQRA to adjust forecast distributions in response to human insights, particularly in settings where domain knowledge complements algorithmic predictions. By explicitly formulating an optimization problem that aligns expert-driven point forecasts with probabilistic distributions, CQRA provides a structured way to incorporate human intuition into data-driven forecasting models, making it well-suited for FPSC and other humanitarian supply chains.}

\textcolor{ForestGreen}{On the other hand, Entropic Tilting has been used as a flexible approach to adjust predictive densities by incorporating external information while maintaining the statistical properties of the original model. This method systematically reweights the probability distribution of forecasts to ensure coherence with expert knowledge or external nowcasts. For example,}
\citet{krüger2017}
\textcolor{ForestGreen}{applied Entropic Tilting to Bayesian VAR forecasts by incorporating short-term nowcasts, while}
\citet{metaxoglou2016}
\textcolor{ForestGreen}{used it to refine option-implied predictive densities for equity returns. These approaches allow for seamless adjustments of predictive distributions while minimizing distortions to the baseline statistical model. While Entropic Tilting provides a suitable method for combining probabilistic and point forecasts, our proposed CQRA model differs in its optimisation framework, which ensures alignment between expert-driven point forecasts and probabilistic forecasts in a constrained manner. Unlike Entropic Tilting, which reweights distributions without necessarily enforcing coherence between expert forecasts and the final probabilistic forecast, CQRA imposes constraints that ensure consistency between the adjusted forecast distribution and expert input. This makes CQRA particularly well-suited for supply chain applications where domain knowledge plays a critical role in refining model-based predictions.}

Let the quantile levels be defined as:

\[
\{q_1, q_2, \dots, q_n\} \quad \text{where} \quad q_i \in [0.01, 0.99]
\]

For a set of weights \({w_1, w_2, \dots, w_n}\) corresponding to these
quantiles, the weighted quantile forecast \(\hat{y_t}^{(q_i)}\) for each
quantile \(q_i\) at time \(t\) is given by:

\[
\hat{y}_t^{(q_i)} = w_i \cdot \text{ProbForecast}_t^{(q_i)}
\]

where \(\text{ProbForecast}_t^{(q_i)}\) is the probabilistic forecast
for quantile \(q_i\) at time \(t\).

The pinball loss for a quantile \(q_i\) and the point forecast
\(\text{PointForecast}_t\) is defined as:

\[
L_{q_i}(y_t, \hat{y}_t^{(q_i)}) = (y_t - \hat{y}_t^{(q_i)}) \cdot \left(q_i - \mathbf{1}(y_t < \hat{y}_t^{(q_i)})\right)
\]

The weighted mean forecast across all quantiles is calculated as:

\[
\bar{y}_t = \frac{1}{n} \sum_{i=1}^{n} \hat{y}_t^{(q_i)}
\]

The total loss \(L_t\) across all quantiles for a single time point
\(t\) is expressed as:

\[
L_t = \sum_{i=1}^{n} \left| \text{PointForecast}_t - \bar{y}_t \right| + \sum_{i=1}^{n} \max\left(0, (\text{PointForecast}_t - \hat{y}_t^{(q_i)}) \cdot \left(q_i - \mathbf{1}(\text{PointForecast}_t < \hat{y}_t^{(q_i)})\right)\right)
\] where;
\(\sum_{i=1}^{n} \left| \text{PointForecast}_t - \bar{y}_t \right|\)
measures the absolute difference between point forecast and weighted
mean quantile forecasts.

The objective is to minimize the total loss \(L_t\) by optimizing the
weights \(w_i\) across all quantiles:

\[
\min_{w_1, w_2, \dots, w_n} \sum_{t=1}^{T} L_t
\]

subject to:

\[
0 \leq w_i \leq 1, \quad \sum_{i=1}^{n} w_i = 1
\]

Once the optimal weights are identified, the final adjusted quantile
forecast \(\tilde{y}_t^{(q_i)}\) for each quantile \(q_i\) is:

\[
\tilde{y}_t^{(q_i)} = w_i^{*} \cdot \text{ProbForecast}_t^{(q_i)}
\]

However, probabilistic forecasts may have a more dominant influence in
this approach since the weights \(w_i\) are restricted to the range
\([0, 1]\). This implies that point forecasts are expected to align
closely with the forecast distribution. When the probabilistic forecasts
are reliable and the point predictions do not significantly deviate from
the mean of the probabilistic forecast, this approach is recommended.
However, recognizing that this is often not the case in practice,
\textcolor{ForestGreen}{we relax the sum-to-one constraint on the combination weights to provide greater flexibility in incorporating expert point forecasts and accounting for bias. A similar relaxation has been explored in prior research (e.g.,}
\citep{granger1984}),
\textcolor{ForestGreen}{which suggests that imposing weight constraints can sometimes lead to suboptimal forecasts by failing to account for systematic biases. Furthermore,}
\citet{trapero2019}
\textcolor{ForestGreen}{also show that an unconstrained approach to optimizing quantile forecast combinations can yield improvements, particularly in applications requiring adaptive weight adjustments. By allowing weights to exceed unity and introducing a bias factor, our modified CQRA approach accommodates situations where expert forecasts systematically deviate from model-based predictions, enhancing the adaptability and robustness of the combined forecast.}

\textbf{\emph{Modified Approach:}}

\begin{enumerate}
\def\labelenumi{\arabic{enumi}.}
\tightlist
\item
  We update the weighted quantile forecast \(\hat{y}_t^{(q_i)}\) as
  follows:
\end{enumerate}

\[
\hat{y}_t^{(q_i)} = w_i \cdot \text{ProbForecast}_t^{(q_i)} + b_t
\]

where \(b_t\) is the bias factor at time \(t\) and it is calculated as
the parameter optimized alongside the weights.

\begin{enumerate}
\def\labelenumi{\arabic{enumi}.}
\setcounter{enumi}{1}
\tightlist
\item
  We remove the normalization constraint \(\sum_{i=1}^{n} w_i = 1\) and
  increase the upper bound of \(w_i\) to 5:
\end{enumerate}

\[
0 \leq w_i \leq 5
\]

Our sensitivity analysis showed that increasing the upper bound beyond 5
did not significantly improve method performance, making 5 an optimal
choice for balancing flexibility and control.

\begin{enumerate}
\def\labelenumi{\arabic{enumi}.}
\setcounter{enumi}{2}
\tightlist
\item
  After optimization, we apply an adjustment factor to ensure that the
  mean of the forecast distribution aligns with the point forecast:
\end{enumerate}

\[
\tilde{y}_t^{(q_i)} = w_i^{*} \cdot \text{ProbForecast}_t^{(q_i)} \cdot adj_t
\]

The adjustment factor \(adj_t\) is defined as:

\[
adj_t =  \frac{\text{PointForecast}_t}{\bar{y}_t}
\]

These adjustments ensure that the combined probabilistic forecast aligns
with the central tendency of the point forecast while still capturing
the uncertainty in the prediction. This approach is particularly useful
when the probabilistic forecast does not include external variables that
cause significant deviations in demand.

After generating the weighted quantile forecast \(\tilde{y}_t^{(q_i)}\)
for each quantile \(q_i\), we create a smooth forecast distribution by
linearly interpolating between the quantile levels:

\[
\tilde{y}_t(x_j) = \tilde{y}_t^{(q_i)} + \left( \frac{x_j - q_i}{q_{i+1} - q_i} \right) \cdot \left( \tilde{y}_t^{(q_{i+1})} - \tilde{y}_t^{(q_i)} \right)
\]

where \(q_i \leq x_j < q_{i+1}\) and \(x_j\) are the interpolation
points.

The final interpolated forecast distribution \(\tilde{Y}_t\) for all
interpolation points \(x_j\) is:

\[
\tilde{Y}_t = \{ \tilde{y}_t(x_1), \tilde{y}_t(x_2), \dots, \tilde{y}_t(x_m) \}
\]

\textbf{Remark 1}: When providing point forecasts to the method, they
should first be combined with the mean forecasts from the probabilistic
forecast using a simple averaging method. This combined point forecast
will serve as the new central tendency (e.g., mean or median) for the
overall forecast. Based on this combined central tendency, optimal
weights will then be determined to enhance the accuracy and balance of
the forecast by integrating both the point and probabilistic
perspectives effectively.

\textbf{Remark 2}: We refer to the first proposed combination method as
the \emph{Hybrid Weighted Average}, and the revised version of the
combination method is termed the \emph{Hybrid Bias Adjustment}.

Underlying assumptions of the method are;

\begin{enumerate}
\def\labelenumi{\arabic{enumi}.}
\tightlist
\item
  The probabilistic forecasts are well-calibrated.
\item
  The point forecast accurately represents the central tendency of the
  future distribution.
\item
  The weights used for each quantile are restricted to non-negative
  values, ensuring that the final forecast distribution remains in the
  range of the original probabilistic forecasts.
\item
  A linear combination of quantile forecasts, point forecasts, and bias
  adjustment is sufficient to represent the true forecast distribution.
\item
  Linear interpolation between quantile levels accurately reflects the
  true underlying distribution.
\end{enumerate}

\section{Experiment setup}\label{sec-experiment}

\subsection{Data collection and
preprocessing}\label{data-collection-and-preprocessing}

\begin{figure}

\centering{

\includegraphics[width=0.8\textwidth,height=\textheight]{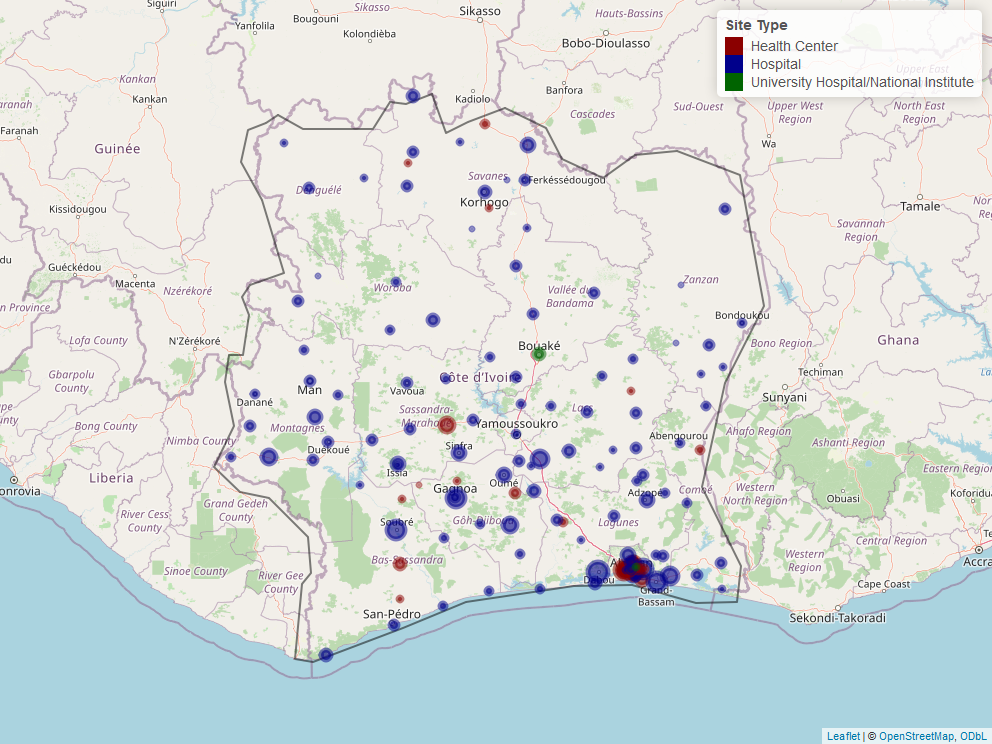}

}

\caption{\label{fig-map}Contraceptive stock distribution in Côte
d'Ivoire by healthcare site location. The size of the circles represents
the quantity of stock distributed.}

\end{figure}

The data used in our study were extracted from the LMIS of Cote
d'Ivoire. The dataset encompassed 156 sites distributed across 81
districts in 20 regions, covering a span of 11 contraceptive products
across 7 product categories. These categories included female and male
condoms, emergency contraceptives, oral contraceptives, injectables,
implants, and IUDs. The dataset spanned from January 2016 to December
2019 at a monthly granularity containing 1454 time series.
Figure~\ref{fig-map} shows the location of each site in Côte d'Ivoire by
site type, illustrating that the sites are distributed throughout the
country.

Our initial exploration indicated that there were no duplicate values;
however, some missing values were present in the time series.
Additionally, a few time series contained stockout cases. Since our
study does not focus on handling stockouts in the forecasting process,
we removed the series with stockouts and missing values, as we could not
determine the reasons behind those missing values. This filtering
resulted in a final dataset of 1,360 time series.

In our study, we focus on stock distributed\footnote{We assume that
  \emph{stock distributed} serves as a reasonable proxy for consumption
  data, as we eliminated stockout cases due to limited access to direct
  consumption data from users.} as the target variable at the site level
for various contraceptive products.

\subsection{Data exploration}\label{data-exploration}

\begin{figure}

\centering{

\includegraphics[width=0.8\textwidth,height=\textheight]{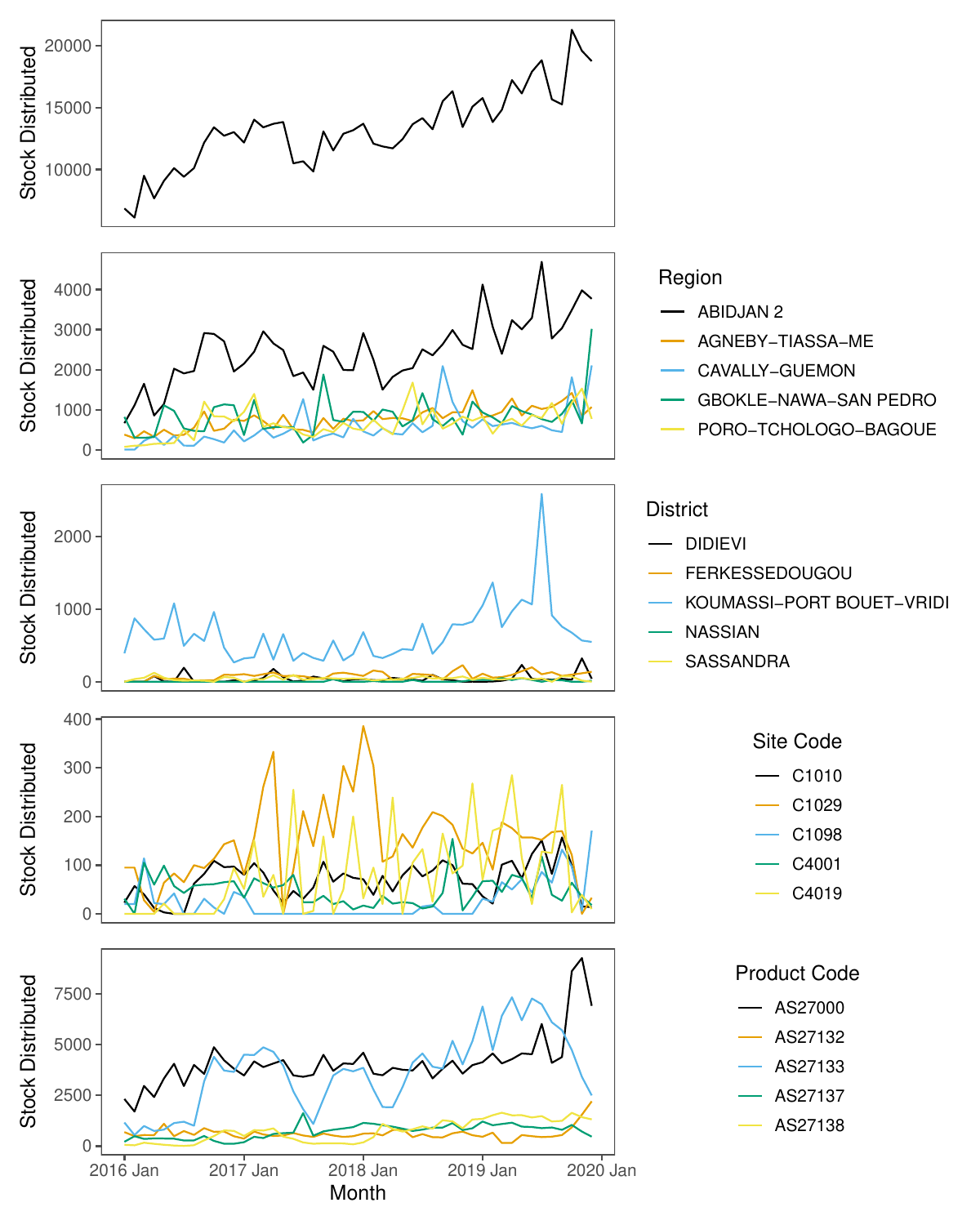}

}

\caption{\label{fig-tsplot1}Time series of contraceptive product stock
distributed (Jan 2016 -- Dec 2019) at various levels. The x-axis
represents the month, while the y-axis indicates the number of units
distributed. The panels display data from the entire country (top
panel), with breakdowns by region, district, site, and product code. The
bottom panel shows the number of units distributed in selected sites for
specific products. To ensure clarity and prevent overplotting, only five
time series are displayed for each aggregate level. These series were
selected randomly and are characteristic of the patterns encountered at
the respective aggregation levels.}

\end{figure}

First, we examined the time plots of the data at various aggregation
levels to observe the time series features such as trend, seasonality,
and noise. As illustrated in Figure~\ref{fig-tsplot1}, higher
aggregation levels reveal clearer seasonal patterns and trends, while
lower aggregation levels exhibit increased volatility. Additionally, the
plots highlight notable differences in stock distribution across
locations and products, suggesting the presence of distinct patterns
associated with each.

\begin{figure}

\centering{

\includegraphics[width=0.8\textwidth,height=\textheight]{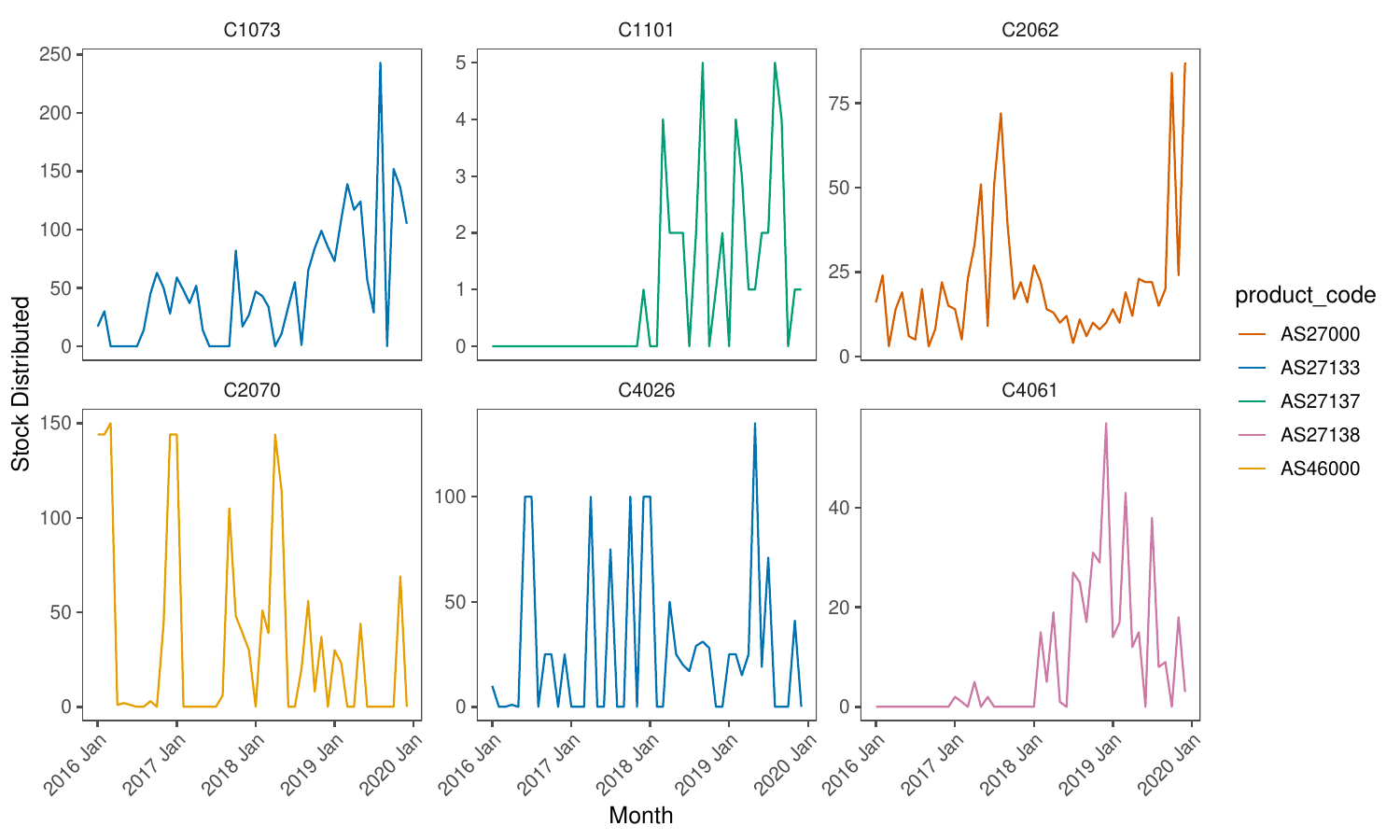}

}

\caption{\label{fig-tsplot2}Time series of contraceptive product stock
distributed in selected sites for specific products(Jan 2016 -- Dec
2019). To ensure clarity and prevent overplotting, only five of the
products are displayed. These series were selected randomly and
represent characteristic patterns at this level.}

\end{figure}

At the lowest aggregation level, depicted in Figure~\ref{fig-tsplot2},
which focuses on product distribution at individual site level, demand
patterns become more heterogeneous, comprising a mix of smooth, erratic,
lumpy, and intermittent demand types. Unlike the aggregate levels, where
trends and seasonality are more apparent, these patterns are not easily
discernible at the site level, further complicating the forecasting
process.

Next, we examined the time series data of the products at the site level
to gain a clearer understanding of trend and seasonality patterns, as
our primary focus is on forecasting each product at the site level.
However, due to the large number of time series, it was not visually
feasible to plot all individual series together to simultaneously
inspect trends and seasonality. Therefore, we employed the Seasonal and
Trend Decomposition using Loess (STL) method \citep{Cleveland1990} to
extract key features from all 1,360 time series.

\begin{figure}

\centering{

\includegraphics[width=0.7\textwidth,height=\textheight]{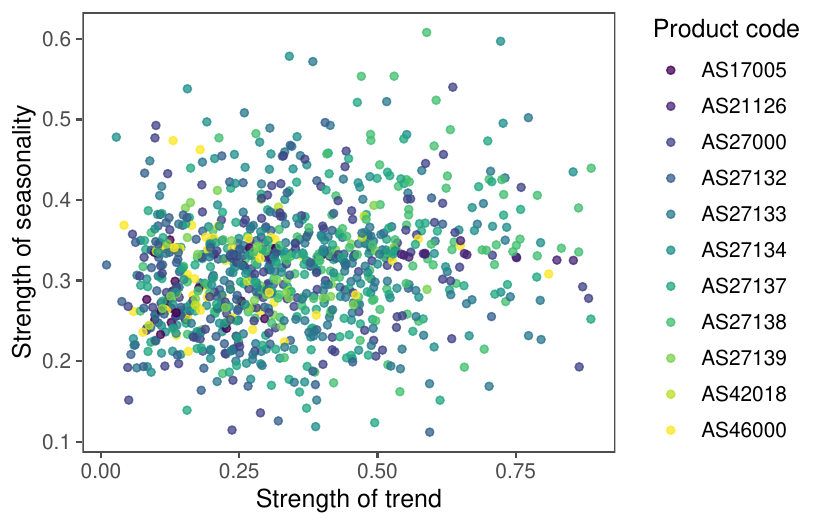}

}

\caption{\label{fig-feature}Trend strength and seasonality in the time
series of stock distribution. Each point in the scatter plot represents
one of the 1,360 time series analysed, with trend and seasonality
strengths measured on a scale from 0 to 1 (0 indicating weak and 1
indicating strong).}

\end{figure}

As shown in Figure~\ref{fig-feature}, the strength of trend and
seasonality for each time series is represented on a scale from 0 to 1,
where 0 indicates low strength and 1 indicates high strength. The
majority of the time series exhibited moderate levels of both trend and
seasonality. However, even within the same product code, we observed
variations in trend and seasonality patterns, which posed challenges for
the forecasting process. Consequently, we considered a range of
forecasting approaches, including time series, Bayesian, ML, and
foundational time series methods, to determine which could most
effectively handle the diverse patterns within the time series.

\subsection{Forecasting setup}\label{forecasting-setup}

Our forecast setup began with data collection and preparation of a tidy
dataset for the forecasting process. Following this, we carried out
feature engineering. As outlined by \citet{kolassa2023}, incorporating
lag predictors and rolling window statistics is beneficial for improving
forecasting methods. In addition to these, we also integrated
categorical and date features into the forecasting process. To ensure we
selected the most relevant variables, a feature importance analysis was
conducted to identify the best predictors for the forecasting methods.

In the USAID forecasting competition, the planning horizon was set to 3
months. However, instead of using fixed training and test sets as in the
competition, we adopted the time series cross-validation approach to
create the training and test sets \citep{hyndman2021forecasting}. Unlike
the fixed approach, where the same training and test sets are used for
evaluation, time series cross-validation moves the forecasting origin
forward by a fixed number of steps, producing multiple forecasts at
different points in time. This allows for the calculation of multi-step
errors, giving a more robust view of how methods perform across various
demand scenarios, such as periods of high and low demand
\citep{svetunkov2023}.

\textcolor{ForestGreen}{In our cross-validation setup, we define the training period as all available data up to September 2019, ensuring sufficient historical observations for model learning. The testing period consists of rolling evaluation windows, where each test set spans a 3-step-ahead forecasting horizon to align with the competition requirements. At each iteration, the training set expands while the test set moves forward by one step, maintaining a consistent evaluation structure.}
We limited the number of rolling origins to 3 per series due to
computational constraints, but this still provided us with meaningful
insights into method performance over time. For forecasting, we employed
recursive multi-step forecasting and we generated 1000 future paths per
a series. All method development and hyper-parameter tuning were
conducted using only the training data to ensure that the evaluation
remained unbiased and the methods were properly validated.

\subsubsection{Probabilistic forecasting using
bootstraping}\label{probabilistic-forecasting-using-bootstraping}

To express the uncertainty of our forecasting methods' estimates, we
utilize probability distributions of potential future values. Several
methods are available to estimate prediction intervals, including
analytical prediction intervals, bootstrapping, quantile regression,
Bayesian modeling (using MCMC sampling), and conformal prediction. In
our study, we employ the bootstrapping method to estimate these
intervals \citep{gneiting2014}.

\textcolor{ForestGreen}{Given that our study employs multiple forecasting methods, bootstrapping provides a unified framework that can be applied consistently without requiring additional parametric assumptions. Moreover, it allows us to approximate the empirical distribution of forecast errors without imposing strict distributional constraints, making it particularly suitable for datasets with varying demand patterns in the FPSC.}

To account for uncertainty in predictions, we assume that future errors
will resemble past errors. We define the error as the difference between
the actual value and the forecasted value:

\[
e_t = y_t-\hat{y}_{t}
\]

where; \(e_t\) is the error at time \(t\), \(y_t\) is the actual value
and \(\hat{y}_{t}\) is the forecast value at time \(t\).

We simulate different future predictions by sampling from the collection
of past errors and adding these to the forecast estimates. Each
bootstrap iteration produces a different potential future path. By
repeating this process, we generate a distribution of possible outcomes.
Based on a chosen significance level, prediction intervals can then be
constructed from this distribution \citep{hyndman2021forecasting}.

To implement the bootstrapping process, we use the fable package in R
for time series methods and the skforecast package for ML methods
\citep{amatrodrigo2023}. However, for the Bayesian methods, this process
is not necessary as it inherently provides probabilistic forecasts as
part of its output. Similarly, foundational time series methods also
deliver probabilistic forecasts directly.

\subsubsection{Forecast combination}\label{forecast-combination}

Forecast combination is a promising approach to enhance forecasting
performance by aggregating multiple forecasts generated using different
methods for a specific time series. This technique eliminates the need
to select a single ``best'' forecasting method, thus leveraging the
strengths of various methods \citep{wang2023}. Known as either
\emph{forecast combination} or \emph{forecast ensemble} across different
fields, this method has been widely used and extensively studied
\citep{godahewa2021}. The literature provides substantial evidence that
forecast combinations consistently outperform individual forecasts
\citep{ranjan}, primarily by mitigating uncertainties arising from data
variability, parameter estimation, and method selection
\citep{wang2023}.

Forecast combination methods can range from linear combinations,
nonlinear combinations, and time-varying weights, to more sophisticated
approaches like cross-learning, correlations among forecasts, or
Bayesian techniques \citep{wang2023}. Among these, the most widely
adopted approach is the linear combination with equal weights
\citep{godahewa2021}. This method is not only straightforward to
implement and interpret but also provides robust and improved
forecasting performance \citep{ranjan, godahewa2021, thompson2024}.
Consequently, in our study, we applied a linear combination approach
with equal weights to generate combined forecasts.

\subsection{Forecasting methods}\label{forecasting-methods}

In our study, we employed a range of forecasting methods to address the
volatile nature of the time series data. For time series methods, we
used sNAIVE, Moving Average (MA), Exponential Smoothing State Space
(ETS), ARIMA, and Syntetos-Boylan approximation (SBA). As Bayesian
methods, we implemented the Bayesian Structural Time Series (BSTS)
method with regressors. For ML methods, we applied Multiple Linear
Regression (MLR), Random Forest (RF), LightGBM (LGBM), and XGBoost
(XGB). These methods were selected due to their popularity, efficiency,
and ease of implementation within the forecasting domain
\citep{makridakis2022}. Furthermore, for the ML methods, we developed
each as a global method, where a single method was trained to produce
forecasts for all time series simultaneously \citep{bandara2021}.

Additionally, we explored foundational time series approaches such as
TimeGPT \citep{garza}, Chronos \citep{ansari} and Lag Llama
\citep{rasul}, which are gaining attention due to advancements in large
language models (LLMs) and also capable of producing probabilistic
forecasts. These methods offer zero-shot forecasting capabilities,
meaning they have been pre-trained on vast amounts of time series data
and can be applied to new time series without the need for retraining or
fine-tuning parameters \citep{carriero}. This feature significantly
reduces the steps typically required in the forecasting process, such as
data preparation, model training, and model selection \citep{garza}.
However, these methodologies have yet to be tested within the FPSC
context.

To offer a more comprehensive comparison of forecasting methods in
contraceptive demand forecasting, we incorporated a demographic
forecasting method. This method uses demographic data such as population
size, age distribution, and other family planning indicators to estimate
future contraceptive demand \citep{akhlaghi2013}. Given that we did not
have access to the final forecasts generated at the site level by demand
planners, we assumed that the demographic-based method serves as a proxy
for expert-driven forecasts. This assumption is grounded in the fact
that experts typically leverage their domain knowledge when determining
key family planning indicators.

\subsubsection{Time series methods}\label{time-series-methods}

\textbf{\emph{sNAIVE}}: This method is a simple forecasting approach
where forecasts are generated using the most recent observation from the
corresponding period of the previous cycle. This method is often used as
a benchmark in time series forecasting \citep{hyndman2021forecasting}
and which can be shown as;

\[
\hat{y}_{t+h}=y_{t+h-s}
\]

Where \(\hat{y}_{t+h}\) is the forecast for \(t+h\), and \(s\) is the
seasonality period. We implemented this method using the SNAIVE()
function in the fable package in R \citep{fable2022}.

\textbf{\emph{MA}}: MA method is a simple forecasting approach that
generates predictions by averaging a fixed number of the most recent
observations. This method helps to smooth out short-term fluctuations
while emphasizing longer-term trends in the data. The method assumes
that future values can be reasonably estimated based on the mean of past
values over a specified window \citep{hyndman2021forecasting}. MA method
can be represented as:

\[
\hat{y}_{t+1} = \frac{1}{n} \sum_{i=0}^{n-1} y_{t-i}
\]

where \(\hat{y}{t+1}\) is the forecast for the next time period, \(n\)
is the number of past observations (the window size), and \(y{t-i}\) are
the actual values from previous periods.Due to the simplicity nature of
this method, it is often used in the FPSC context. We implemented this
method using the MEAN() function in the fable package in R
\citep{fable2022}

\textbf{\emph{ETS}}: ETS model accommodates trends, seasonality, and
error terms within time series data through various approaches, such as
additive, multiplicative, or mixed models within a state-space
framework. The model updates these components dynamically over time
using recursive equations. The ETS model is capable of handling diverse
time series patterns, including trends and seasonal fluctuations
\citep{hyndman2021forecasting}. Given the large number of series in our
dataset, we utilized the automated ETS model, which selects the optimal
model based on Akaike's Information Criterion (AIC) for each time
series. We used the ETS() function in the fable package in R
\citep{fable2022} to implement this model.

\textbf{\emph{ARIMA}}: ARIMA model forecasts based on trends,
autocorrelation, and noise within time series data. It is also flexible
and can handle both non-seasonal and seasonal data by incorporating
seasonal components. ARIMA parameters (\emph{p,d,q}) denote the orders
of the auto-regressive (AR) component, differencing, and moving average
(MA) component, respectively. ARIMA is particularly effective for data
with a pronounced temporal structure \citep{hyndman2021forecasting}.
Like with the ETS model, we employed an automated approach to fit ARIMA
models for each time series using the ARIMA() function in the fable
package in R, which selects the best model using similar criteria
\citep{fable2022}.

\textbf{\emph{SBA}}: Since some time series exhibit an intermittent
demand nature, we also employed the SBA method in our study, an
enhancement of Croston's original method from 1972 \citep{syntetos2005}.
The SBA approach methods intermittent demand as a binomial process by
separately estimating the demand intervals and the demand sizes when
they occur. This method applies a correction factor to reduce the
inherent positive bias of the original Croston method, making the
forecasts more accurate. We implemented this method using the
CROSTON(type = `sba') function in the fable package in R
\citep{fable2022}.

\subsubsection{Bayesian methods}\label{bayesian-methods}

\textbf{\emph{BSTS}}: The BSTS method used in our study combines a local
linear trend and a seasonal component, incorporating additional
covariates to fit the observed data \citep{kohns2023}. The local linear
trend is a time-varying method that captures the evolving pattern of the
time series over time. It consists of a level and a slope, both of which
are allowed to change dynamically. The state equations for this are:

\[
\mu_t = \mu_{t-1} + \beta_{t-1} + \eta_t, \quad \eta_t \sim N(0, \sigma_{\eta}^2)
\] \[
\beta_t = \beta_{t-1} + \zeta_t, \quad \zeta_t \sim N(0, \sigma_{\zeta}^2)
\] where; \(\mu_t\) is the level at \(t\), \(\beta_t\) is the slope at
\(t\), \(\eta_t\) and \(\zeta_t\) are normally distributed errors with
variances \(\sigma_{\eta}^2\) and \(\sigma_{\zeta}^2\) respectively.

The seasonal component captures regular patterns that repeat over a
fixed period and it is modeled as:

\[
S_t = -\sum_{j=1}^{m-1} S_{t-j} + \omega_t, \quad \omega_t \sim N(0, \sigma_{\omega}^2)
\]where; \(S_t\) is the seasonal effect at time \(t\), \(m\) is the
number of seasons (in our case, \(m\) = 12 for monthly data) and \(w_t\)
is the normally distributed error with variance \(\sigma_{\omega}^2\).

The observed data (i.e., target variable) is modeled as a linear
combination of the local linear trend, seasonal component, and
additional regressors. This is represented by the observation equation:;

\[
y_t = \mu_t + S_t + X_t \beta + \epsilon_t, \quad \epsilon_t \sim N(0, \sigma_{\epsilon}^2)
\]

where; \(X_t\) are the regressors, \(\beta\) are the corresponding
coefficients and \(\epsilon_t\) is the noise.

Posterior distributions for parameters are estimated using Markov Chain
Monte Carlo (MCMC) methods \citep{kohns2023}. The predicted future
values are obtained by simulating from these posterior distributions and
thus quantifying the uncertainty given the Bayesian nature of the method
\citep{martin2024}.

\subsubsection{ML methods}\label{ml-methods}

\textbf{\emph{MLR}}: MLR methods establish linear relationships between
the target variable and multiple predictor variables. The method
estimates coefficients for each predictor variable by minimizing the
residual sum of squares between observed and predicted values. These
methods are particularly useful when demand is influenced by various
factors \citep{hyndman2021forecasting}. In our study, we implemented
this method using the LinearRegression() function from the sklearn
package in Python \citep{scikit-learn}.

\textbf{\emph{RF}}: RF is an ensemble learning method that constructs a
collection of decision trees, each trained on a bootstrap sample of the
original data. The predictions of these trees are aggregated to produce
the final forecast \citep{breiman2001}. We used the
RandomForestRegressor() function from the sklearn package in Python
\citep{scikit-learn} to implement the RF method.

\textbf{\emph{Gradient Boosted Regression Trees (LGBM and XGB)}}: These
methods are known for their efficiency and ease of implementation
\citep{makridakis2022}. These methods use an ensemble of decision trees,
where each new tree is added to correct the residuals of the previous
trees in an iterative manner \citep{januschowski2022}. Unlike Random
Forest, which builds trees independently, gradient boosting methods
focus on improving method performance iteratively. In our study, we
selected LightGBM (LGBM) and XGBoost (XGB) for their ability to handle
multiple predictor variables in various forms (binary, categorical, and
numeric) and their effectiveness in providing reliable and accurate
forecasts \citep{makridakis2022}. We used the LGBMRegressor() function
from the LightGBM package in Python \citep{MicrosoftCorporation2022} and
the XGBRegressor() function from the XGBoost package in Python
\citep{XgboostDevelopers2021}. Hyperparameter tuning for both LGBM and
XGB was performed using grid search, with the Poisson distribution
chosen as the objective function due to the count nature of the target
variable.

\subsubsection{Foundational time series
methods}\label{foundational-time-series-methods}

\textbf{\emph{TimeGPT}}: TimeGPT is the first pre-trained foundational
method specifically designed for time series forecasting, developed by
Nixtla \citep{garza}. It employs a transformer-based architecture with
an encoder-decoder setup, but unlike other methods, it is not derived
from existing large language methods (LLMs); rather, it is purpose-built
to handle time series data. TimeGPT was trained on over 100 billion data
points, encompassing publicly available time series from a variety of
domains, including retail, healthcare, transportation, demographics,
energy, banking, and web traffic. Due to the diversity of these data
sources and the range of temporal patterns they exhibit, TimeGPT can
effectively handle a wide variety of time series characteristics.
Additionally, the method can incorporate external regressors into the
forecasting process and is capable of producing quantile forecasts,
allowing for robust uncertainty estimation (see \citet{ansari} for a
detailed overview).

\textbf{\emph{Chronos}}: Chronos is a univariate probabilistic
foundational time series method developed by Amazon \citep{ansari}. Like
TimeGPT, it is based on a transformer architecture in an encoder-decoder
configuration, but it trains an existing LLM architecture using
tokenized time series via cross-entropy loss. Chronos was pre-trained on
a large publicly available time series dataset, as well as on simulated
data generated through Gaussian processes. The method was trained on 28
datasets, comprising approximately 84 billion observations. Chronos is
based on the T5 family of methods, offering different versions with
parameter sizes ranging from 20 million to 710 million. The four
pre-trained methods available for forecasting are: 1) Mini (20 million),
2) Small (46 million), 3) Base (200 million), and 4) Large (710 million)
(see \citet{ansari} for a detailed overview). In our study, we employed
the Base Chronos T5 method for its balance between performance and
computational efficiency.

\textbf{\emph{Lag Llama}}: Lag Llama is another univariate probabilistic
foundational time series method, which is based on the LLaMA
architecture and utilizes a decoder-only structure \citep{rasul}. The
method tokenizes time series data using lags as covariates and applies
z-normalization at the window level. This approach focuses on learning
time series behavior from past observations. Lag Llama was trained on 27
publicly available time series datasets across six domains: nature,
transportation, energy, economics, cloud operations, and air quality.
With 25 million parameters, this method is designed to handle diverse
time series frequencies and features, making it suitable for a wide
range of forecasting tasks (see \citet{rasul} for a detailed overview).

\subsubsection{Demographic forecasting
method}\label{demographic-forecasting-method}

In the FPSC context, the demographic forecasting method is employed to
estimate contraceptive needs for a given population based on a set of
family planning indicators during the forecast period. This method is
formulated as a combination of these indicators and population dynamics,
as represented in the following equation \citep{akhlaghi2013}:

\[
y_{i,t} = \left( \sum_{j=15}^{50} \left(mCPR_{t,j} \times Women Population_{t,j} \right) \right) \times Method Mix_{t,i} \times CYP_{t,i} \times Brand Mix_{t,i} \times Source Share_t
\]

Where \(i\) represents the product, \(j\) is the age group, \(mCPR\) is
the modern contraceptive prevalence rate, and \(CYP\) refers to
couple-years of protection.

\(Women Population_{t,j}\) denotes the total population of women in a
selected location, typically within the age range of 15-49 years, which
is the standard range used in census data or demographic health surveys.
For our study, we sourced this population data from WorldPop
\citep{worldpop} and mapped it to each healthcare site based on the
latitude and longitude coordinates of those sites.

\(mCPR\) stands for the percentage of women of reproductive age using
modern contraceptives, with data collected from the PMA Data Lab
\citep{pmadata}.

\(Method Mix\) represents the share of different contraceptive methods
being used, including injectables, IUDs, implants, pills, and condoms.
This data is also obtained from the PMA Data Lab \citep{pmadata}.

\(CYP\) is a metric estimating the protection from pregnancy provided by
a contraceptive method over one year. For example, an implant can cover
3.8 years, so CYP adjusts for such longer-acting methods. We collected
this data from USAID \citep{usaid_cyp}.

\(Brand Mix\) reflects the brand share percentage within each
contraceptive method. This was calculated using historical data.

\(Source Share\) refers to where women of reproductive age, using a
specific method and brand, obtain their products. This mix typically
includes public, private, NGO/SMO (social marketing organizations), and
other small providers. Data was gathered through discussions with USAID
officials.

This equation provides \(y_{i,t}\), which is the total annual point
estimate of contraceptives required for product \(i\) at time \(t\). It
is typically used at the national level on an annual basis to inform
procurement decisions \citep{akhlaghi2013}.

However, as our study focuses on monthly estimates at the healthcare
site level, we revised the equation by introducing a weighting factor,
\(w_t\), to distribute the annual estimates across months. The revised
equation is as follows:

\[
y_{i,t,s} = \left( \sum_{j=15}^{50} \left(mCPR_{t,j} \times Women Population_{t,j,s} \right) \right) \times Method Mix_{t,i} \times CYP_{t,i} \times Brand Mix_{t,i} \times \ Source Share_t \times w_t
\]

Where \(w_t\) represents the monthly weight, \(s\) is the healthcare
site, and \(y_{i,t,s}\) is the monthly point forecast for product \(i\)
at healthcare site \(s\).

\subsubsection{Overview of candidate
methods}\label{overview-of-candidate-methods}

In our study, we developed 20 candidate methods by experimenting with
different combinations of predictors and by combining various
forecasting methods. For the MA method, we opted to use a three-month
averaging period, aligning with the current practice at the site level
in Côte d'Ivoire. Additionally, we developed two model combinations
using equal-weight linear averaging: a combined statistical model and a
combined ML model.

To create hybrid probabilistic methods, we combined point forecasts from
the demographic method with the combined ML method, resulting in a
hybrid combined method. This hybrid method synthesizes insights from
both the demographic point forecast and the probabilistic
algorithm-based forecast, aiming to capture expert knowledge alongside
data-driven characteristics of machine learning methods. This
integration is intended to enhance forecast accuracy by leveraging the
strengths of both approaches.

We developed two variations of the hybrid probabilistic method based on
our proposed methods. A detailed overview of all 20 candidate methods is
provided in Table~\ref{tbl-models}. We also explored several other
approaches to develop different forecast method variations. These
included using demographic indicators as predictors, different method
combinations and applying hierarchical forecasting reconciliation to
combine demographic-based forecasts with algorithm-based forecasting
methods. However, we decided not to include the results of these
methods, as they did not improve the performance significantly.

\begin{landscape}\begingroup\fontsize{8}{10}\selectfont

\begin{longtable}[t]{>{\raggedright\arraybackslash}p{6em}>{\raggedright\arraybackslash}p{6em}>{\raggedright\arraybackslash}p{28em}>{\raggedright\arraybackslash}p{28em}>{\raggedright\arraybackslash}p{6em}}

\caption{\label{tbl-models}Proposed candidate methods in our study}

\tabularnewline

\toprule
Type & Method & Predictor variables & Remarks & Probabilistic Forecasts\\
\midrule
Time series & sNaive & Historical stock distributed data & - & Yes\\
 & Moving average & Historical stock distributed data & - & Yes\\
 & ETS & Historical stock distributed data & - & Yes\\
 & ARIMA & Historical stock distributed data & - & Yes\\
 & Croston-SBA & Historical stock distributed data & - & No\\
\addlinespace
Bayesian & BSTS reg & Historical stock distributed data, lag values (for 1,2,3,4), lag rolling mean, 4 period rolling max, 4 period rolling zero percentage, Month and year, region, district, site type, site code, product type and product code & - & Yes\\
 & BSTS demo & Historical stock distributed data, women population at each site, mCPR, method mix,CYP,  brand mix, source share & - & Yes\\
ML & MLR & Historical stock distributed data, lag values (for 1,2,3,4), lag rolling mean, 4 period rolling max, 4 period rolling zero percentage, Month and year, region, district, site type, site code, product type and product code & - & Yes\\
 & RF &  & - & Yes\\
 & LGBM &  & - & Yes\\
\addlinespace
 & XGB &  & - & Yes\\
Demographic & Demographic & Women population at each site, mCPR, method mix,CYP,  brand mix, source share, weight for each month & - & No\\
Foundational & TimeGPT & Historical stock distributed data & - & Yes\\
 & TimeGPT reg & Historical stock distributed data, lag values (for 1,2,3,4), lag rolling mean, 4 period rolling max, 4 period rolling zero percentage, Month and year, region, district, site type, site code, product type and product code & - & Yes\\
 & Chronos & Historical stock distributed data & - & Yes\\
\addlinespace
 & Lag Llama & Historical stock distributed data & - & Yes\\
Combination & Statistical combined & - & Model combination using sNAVIE, MA, ETS, ARIMA. We didn’t use Croston as it only produces point forecats. & Yes\\
 & ML combined & - & Model combination using RF, LGBM, XGB. We didn't use MLR because it significantly reduces combined forecast performance. & Yes\\
Hybrid & Hybrid weighted average & - & Combination between demographic mdethod and ML combination using the weighted average approach. & Yes\\
 & Hybrid bias adjustment & - & Combination between demographic method and ML combination using the weighted average bias approach. & Yes\\
\bottomrule

\end{longtable}

\endgroup{}
\end{landscape}

\subsection{Performance evaluation}\label{performance-evaluation}

To assess the performance of our forecasting methods, we used both point
forecast and probabilistic forecast evaluation metrics. We evaluated
point forecasts using the MASE.
\textcolor{ForestGreen}{MASE was chosen for two primary reasons. First, it was the official evaluation metric used in the USAID Intelligent Forecasting Competition, allowing us to directly compare our model performance with previous benchmark results. Second, MASE is a scale-independent metric that provides robustness, and stability}
\citep{kolassa2023}.

The MASE formula is:

\[
  \text{MASE} = \text{mean}(|q_{t}|),
\]

where

\[
  q_{t} = \frac{ e_{t}}
 {\displaystyle\frac{1}{n-m}\sum_{t=m+1}^n |y_{t}-y_{t-m}|},
\]

Here, \(e_t\) is the point forecast error for forecast period \(t\),
\(m=12\) (to account for seasonality), \(y_t\) is the observed value,
and \(n\) is the number of observations in the training set. The
denominator is the mean absolute error of the seasonal naive method over
the training sample, ensuring the error is properly scaled. Smaller MASE
values indicate more accurate forecasts, and since it was the metric
used in the USAID competition, it allows us to compare our results with
the competition submissions.

To evaluate the accuracy of probabilistic forecasts, we employed the
CRPS, a widely used metric in probabilistic forecasting that assesses
the sharpness and calibration of the forecast distribution.

The CRPS is given by:

\[
  \text{CRPS} = \text{mean}(p_j),
\]

where

\[
  p_t = \int_{-\infty}^{\infty} \left(G_t(x) - F_t(x)\right)^2dx,
\]

where \(G_t(x)\) is the forecasted probability distribution function for
the period \(t\), and \(F_t(x)\) is the true probability distribution
function for the same period.

CRPS is beneficial to our study as it measures the overall performance
of the forecast distribution by rewarding sharpness and penalizing
miscalibration \citep{gneiting2014}. Calibration measures how well
predicted probabilities match the true observations, while sharpness
focuses on the concentration of the forecast distributions
\citep{wang2023}. Thus, CRPS provides a single score by evaluating both
calibration and sharpness, making it easy to evaluate the performance of
forecasting methods. In this formula, \(G_t(x)\) is the forecasted
cumulative distribution function (CDF) for time \(t\) and \(F_t(x)\) is
the true CDF for the same time. The CRPS evaluates the difference
between the predicted and actual probability distributions, with lower
values indicating better performance \citep{ranjan}. It combines aspects
of both calibration (the alignment of predicted probabilities with
actual outcomes) and sharpness (the concentration of the forecast
distribution), making it a comprehensive measure of forecast quality
\citep{wang2023}.

\textcolor{ForestGreen}{While CRPS provides a comprehensive evaluation of the entire predictive distribution, there are cases where accuracy at specific quantiles is of particular interest. For example, in inventory management of family planning health commodities, higher quantiles (e.g., the 95th percentile) might be important to ensure efficient stock management and maintain a high service level. By accurately capturing demand at these upper quantiles, supply chain planners can better mitigate stockouts and ensure the consistent availability of essential health supplies. In such scenarios, the quantile score (or pinball score) becomes a more appropriate metric, as it directly evaluates forecast accuracy at the chosen quantile, ensuring a more targeted assessment of predictive performance.}

\section{Analysis and results}\label{sec-results}

First, we evaluate the overall point forecast performance of the
forecasting methods, including the proposed method, using the MASE.
Additionally, we compare the overall performance of our methods against
the top 10 submissions from the USAID competition. Second, we assess the
overall performance of the probabilistic forecasts of our methods using
the CRPS. After completing these evaluations, we conduct a Nemenyi test
at the 5\% significance level to determine any significant differences
in performance between the methods.

Next, we evaluate both the point and probabilistic forecast performances
across forecast horizons, providing a clearer picture of multi-step
errors in the methods. Following this, we compare the forecast
performances in relation to computational time, highlighting the
trade-offs between accuracy and efficiency.

\subsection[Overal performance evaluation of point and probabilistic
forecasts]{\texorpdfstring{Overal performance evaluation of point and
probabilistic
forecasts\footnote{Overall performance refers to the mean and median
  forecast performance of methods calculated on the test sets at
  forecast horizons \(h=1,2,3\) months, with time series
  cross-validation applied to the target variable.}}{Overal performance evaluation of point and probabilistic forecasts}}\label{overal-performance-evaluation-of-point-and-probabilistic-forecasts04-results-1}

The overall point forecast performance of each method is presented in
Table~\ref{tbl-mase}, showing both mean and median MASE values, and
ordered by mean MASE. The table clearly indicates that all time series
methods underperform compared to ML methods. In fact, the top five
methods are ML-based. The top-performing method is the RF method, with
the lowest mean MASE of 0.743. Notably, the Hybrid Weighted Average
method is the second-best performer, with a mean MASE of 0.775.

However, the Hybrid Bias Adjustment method performs significantly worse
compared to all other methods, except for the Demographic method, which
shows the poorest performance among all methods. Interestingly, the SBA
method outperforms all other time series methods, but neither the
Statistical Combined method nor the ML Combined method surpass other
methods within their respective categories as initially expected.
Nevertheless, it is notable that both combined methods improve their
performance compared to the lowest-performing methods within their
category.

On the other hand, the BSTS method shows improved performance when it
incorporates time series-based predictors (e.g., lags, rolling
statistics), categorical features (e.g., region, district), and date
features, compared to when it uses demographic-based predictors (e.g.,
women population, mCPR, women age group). Among the foundational
methods, TimeGPT with regressors outperforms all other foundational
methods, whereas without regressors, the performance of Chronos and
TimeGPT is quite similar. However, the performance of Lag Llama differs
notably from both Chronos and TimeGPT. Finally, regarding the
competition submissions, none of them outperform the top five methods in
our analysis.

\begin{table}

\caption{\label{tbl-mase}Overall point forecast accuracy in mean MASE
and median MASE (CS refers to competition submission).}

\centering{

\centering\begingroup\fontsize{10}{12}\selectfont

\resizebox{\ifdim\width>\linewidth\linewidth\else\width\fi}{!}{
\begin{tabular}{>{\raggedright\arraybackslash}p{15em}>{\raggedleft\arraybackslash}p{8em}>{\raggedleft\arraybackslash}p{8em}}
\toprule
Method & Mean MASE & Median MASE\\
\midrule
RF & 0.743 & 0.376\\
Hybrid weighted averaging & 0.775 & 0.426\\
LGBM & 0.833 & 0.426\\
ML combined & 0.847 & 0.460\\
XGB & 0.859 & 0.433\\
CS 01 & 0.990 & 0.789\\
CS 02 & 0.995 & 0.798\\
CS 03 & 0.998 & 0.779\\
CS 04 & 1.004 & 0.790\\
CS 05 & 1.014 & 0.815\\
CS 06 & 1.035 & 0.785\\
CS 07 & 1.043 & 0.823\\
CS 08 & 1.051 & 0.819\\
CS 09 & 1.088 & 0.844\\
CS 10 & 1.103 & 0.861\\
TimeGPT reg & 1.258 & 0.623\\
MLR & 1.269 & 0.632\\
TimeGPT & 1.292 & 0.669\\
Chronos & 1.305 & 0.641\\
BSTS reg & 1.327 & 0.732\\
SBA & 1.331 & 0.689\\
Moving average & 1.373 & 0.694\\
Statistical combined & 1.378 & 0.731\\
ETS & 1.379 & 0.683\\
ARIMA & 1.386 & 0.689\\
Lag Llama & 1.483 & 0.777\\
BSTS demo & 1.521 & 0.859\\
sNaive & 1.603 & 0.924\\
Hybrid bias adjustment & 4.360 & 0.800\\
Demographic & 16.072 & 1.847\\
\bottomrule
\end{tabular}}
\endgroup{}

}

\end{table}%

\begin{figure}

\centering{

\includegraphics[width=0.8\textwidth,height=\textheight]{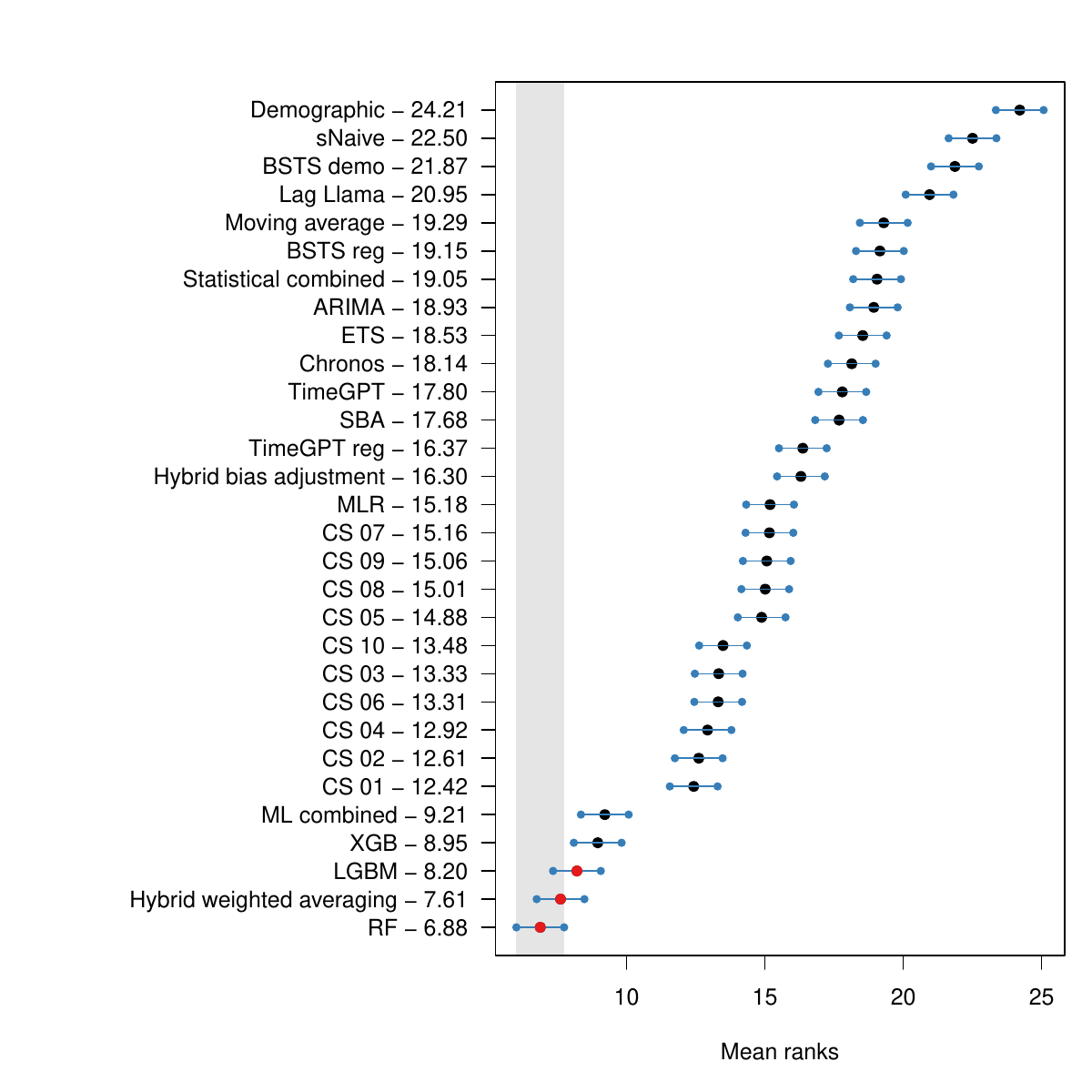}

}

\caption{\label{fig-mase_nemenyi}Average ranks of forecasting methods
with 95\% confidence intervals based on the Nemenyi test for MASE
values. Lower ranks indicate better forecast performance.}

\end{figure}

However, we cannot draw concrete conclusions about the point forecast
performance of methods solely based on mean MASE values. Therefore, we
also conducted the Nemenyi test at the 5\% significance level on MASE
values for the forecasting methods. This test allowed us to calculate
the average ranks of the forecasting methods and assess whether their
performances are significantly different from one another.
Figure~\ref{fig-mase_nemenyi} shows the results of the Nemenyi test.

In brief, if there is no overlap in the confidence intervals between two
methods, it indicates that their performances are significantly
different. The grey area represents the 95\% confidence interval for the
top-ranking method. Methods whose intervals do not overlap with this
grey area are considered significantly underperforming compared to the
best-performing method, and vice versa.

Figure~\ref{fig-mase_nemenyi} demonstrates that the RF method is the
best-performing method confirming our previous finding, and there is no
significant difference between the top three ranked methods, which
include our proposed Hybrid Weighted Average method and the LGBM method.
It is noteworthy that the average rank of the Hybrid Bias Adjustment
method has improved, suggesting that it may perform adequately across a
majority of the time series. Additionally, it is significant that the
TimeGPT with Regressors method outperforms all other foundational time
series methods, which were trained as univariate methods.

Next, we turn our attention to evaluating the performance of
probabilistic forecasts. Table~\ref{tbl-crps} presents the overall
performance evaluations of probabilistic forecasts using both the mean
and median CRPS values, ordered by mean CRPS. The proposed Hybrid
Weighted Averaging method is the top performer, with a mean CRPS of
9.868. The RF method ranks second, with a mean CRPS of 9.997. As in the
point forecast analysis, all the top five methods are ML based, and the
time series methods generally underperform in comparison.

In the BSTS method, we again observe improved performance when time
series-based, categorical, and date features are included as regressors.
The Statistical Combined and ML Combined methods show performance
similar to what was seen in the point forecast analysis.

Notably, Chronos performs better than all time series, Bayesian, and
other foundational methods. Moreover, ETS outperforms all time series
methods but shows poor performance compared to ML based methods. Lastly,
the Hybrid Bias Adjustment method delivers the worst performance among
all forecasting methods, reinforcing the trend observed in the point
forecast evaluation.

\begingroup\fontsize{10}{12}\selectfont

\begin{longtable}[t]{>{\raggedright\arraybackslash}p{15em}>{\raggedleft\arraybackslash}p{8em}>{\raggedleft\arraybackslash}p{8em}}

\caption{\label{tbl-crps}Overall probabilistic forecast accuracy in mean
CRPS and median CRPS.}

\tabularnewline

\toprule
Method & Mean CRPS & Median CRPS\\
\midrule
Hybrid weighted averaging & 9.868 & 3.083\\
RF & 9.997 & 2.754\\
LGBM & 10.131 & 3.067\\
ML combined & 10.286 & 3.377\\
XGB & 10.560 & 3.164\\
\addlinespace
MLR & 12.611 & 4.512\\
Chronos & 15.018 & 4.698\\
BSTS reg & 15.342 & 5.275\\
ETS & 15.397 & 5.632\\
TimeGPT reg & 15.635 & 4.783\\
\addlinespace
Moving average & 15.701 & 5.480\\
ARIMA & 15.703 & 5.602\\
TimeGPT & 15.831 & 5.275\\
Lag Llama & 15.840 & 5.919\\
Statistical combined & 16.045 & 5.671\\
\addlinespace
BSTS demo & 17.064 & 6.526\\
sNaive & 17.511 & 7.119\\
Hybrid bias adjustment & 29.062 & 6.447\\
\bottomrule

\end{longtable}

\endgroup{}

\begin{figure}

\centering{

\includegraphics[width=0.8\textwidth,height=\textheight]{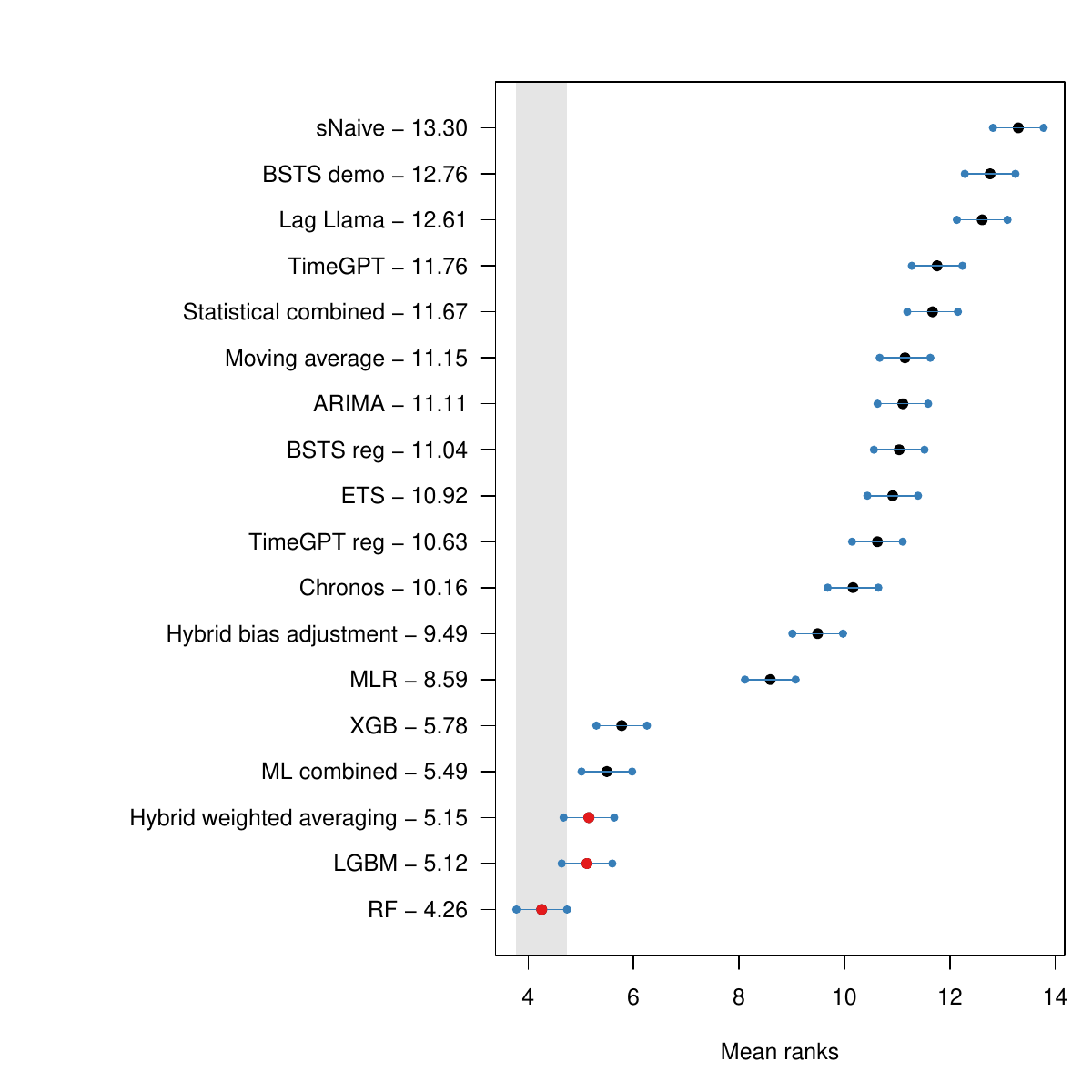}

}

\caption{\label{fig-crps_nemenyi}Average ranks of forecasting methods
with 95\% confidence intervals based on the Nemenyi test for CRPS
values. Lower ranks indicate better forecast performance.}

\end{figure}

Similar to the point forecast analysis, Figure~\ref{fig-crps_nemenyi}
demonstrates that the top three ranked methods are not significantly
different, with RF as the top-ranked method, although the proposed
Hybrid Weighted Average method has the lowest mean CRPS. This may
indicate that RF performs comparably in minimizing the loss function
across series, while the Hybrid Weighted Average method may prioritize
stable time series without significant deviations (see
Section~\ref{sec-model}). Additionally, the top three ranked methods
significantly outperform all other forecasting methods in terms of
probabilistic forecasting.

Noticeably, the Hybrid Bias Adjustment method shows a significant
improvement in its average rank, ranking seventh, right after the ML and
Hybrid Weighted Averaging methods. The Chronos method is also ranked
higher than the time series, BSTS, and other foundational methods. Time
series methods remain clustered in the lower rank range, while the BSTS
reg method shows an improvement in rank compared to the BSTS demo
method. Among the foundational methods, Lag Llama has the lowest rank,
further confirming its relatively weak performance compared to other
foundational and ML based methods.

\subsection{Point and probabilistic forecast performances across
forecast
horizons}\label{point-and-probabilistic-forecast-performances-across-forecast-horizons}

We also analyze the forecast performances over different horizons to
evaluate how the methods perform over time. The forecast horizons range
from month 1 to month 3, corresponding to the upcoming planning period
used by planners for decision-making. First, we examine the error
distribution across all methods. The RF method consistently shows the
highest point forecast accuracy across all three horizons. Additionally,
the top five methods, including the proposed hybrid weighted averaging
method, maintain consistent performance throughout the forecast periods.

In terms of probabilistic forecast accuracy, similar patterns are
observed across different methods. While these plots offer a high-level
overview of error metric distributions (see
Figure~\ref{fig-mase_boxplot} in \textbf{Appendix 1}), they do not
provide detailed insights into the differences between the top- and
low-ranking methods. To gain a clearer understanding of the error
metrics distribution, we plot density distributions, focusing on the top
three and bottom three forecasting methods for both point forecasts and
probabilistic forecasts.

\begin{figure}

\centering{

\includegraphics[width=0.8\textwidth,height=\textheight]{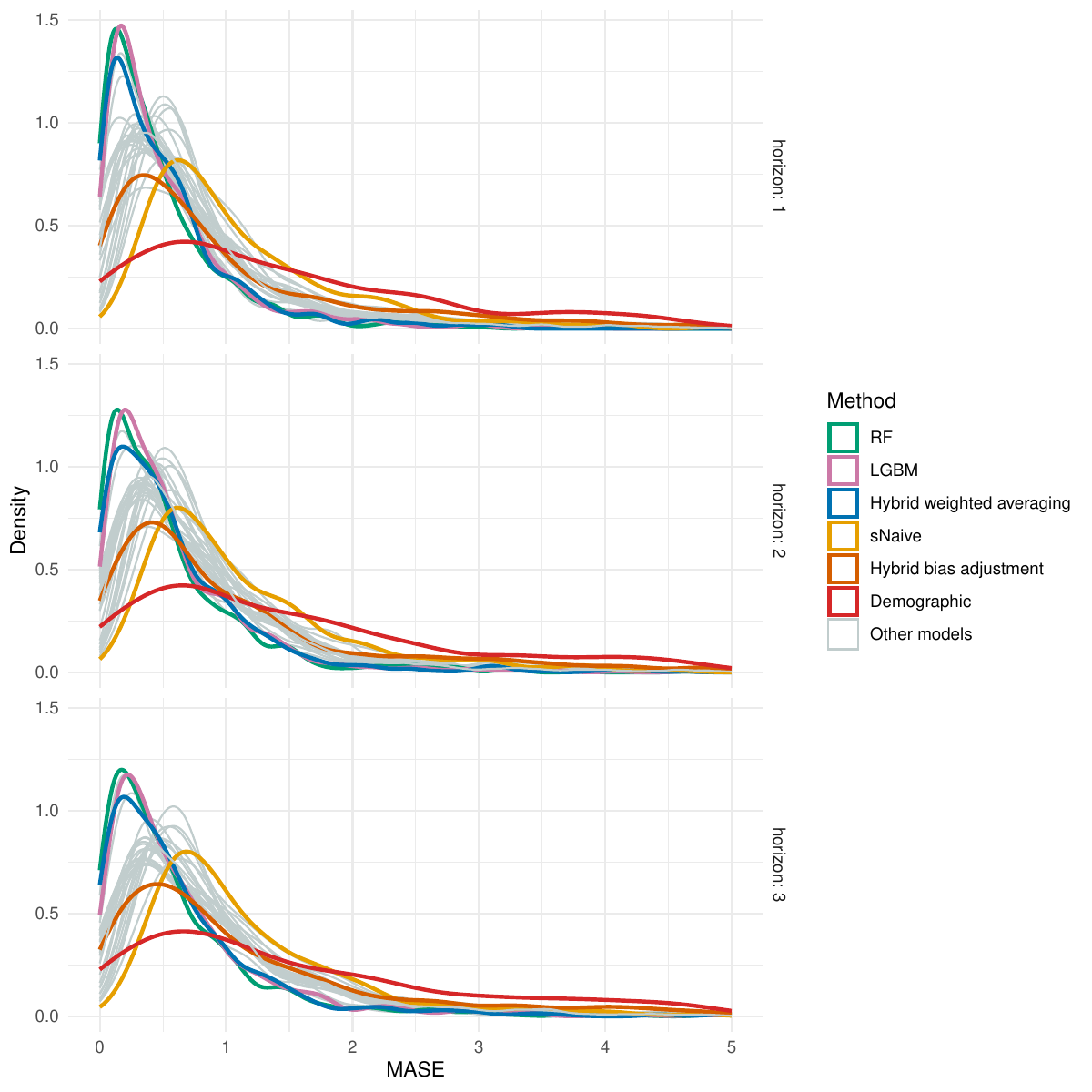}

}

\caption{\label{fig-mase_density}The distribution of MASE values for the
top three and bottom three forecasting methods across the horizons is
presented. The methods are ranked based on their mean MASE values, with
the top and bottom methods selected accordingly. Grey lines represent
the distribution of MASE values for all other methods, providing a
comparative context.}

\end{figure}

\begin{figure}

\centering{

\includegraphics[width=0.8\textwidth,height=\textheight]{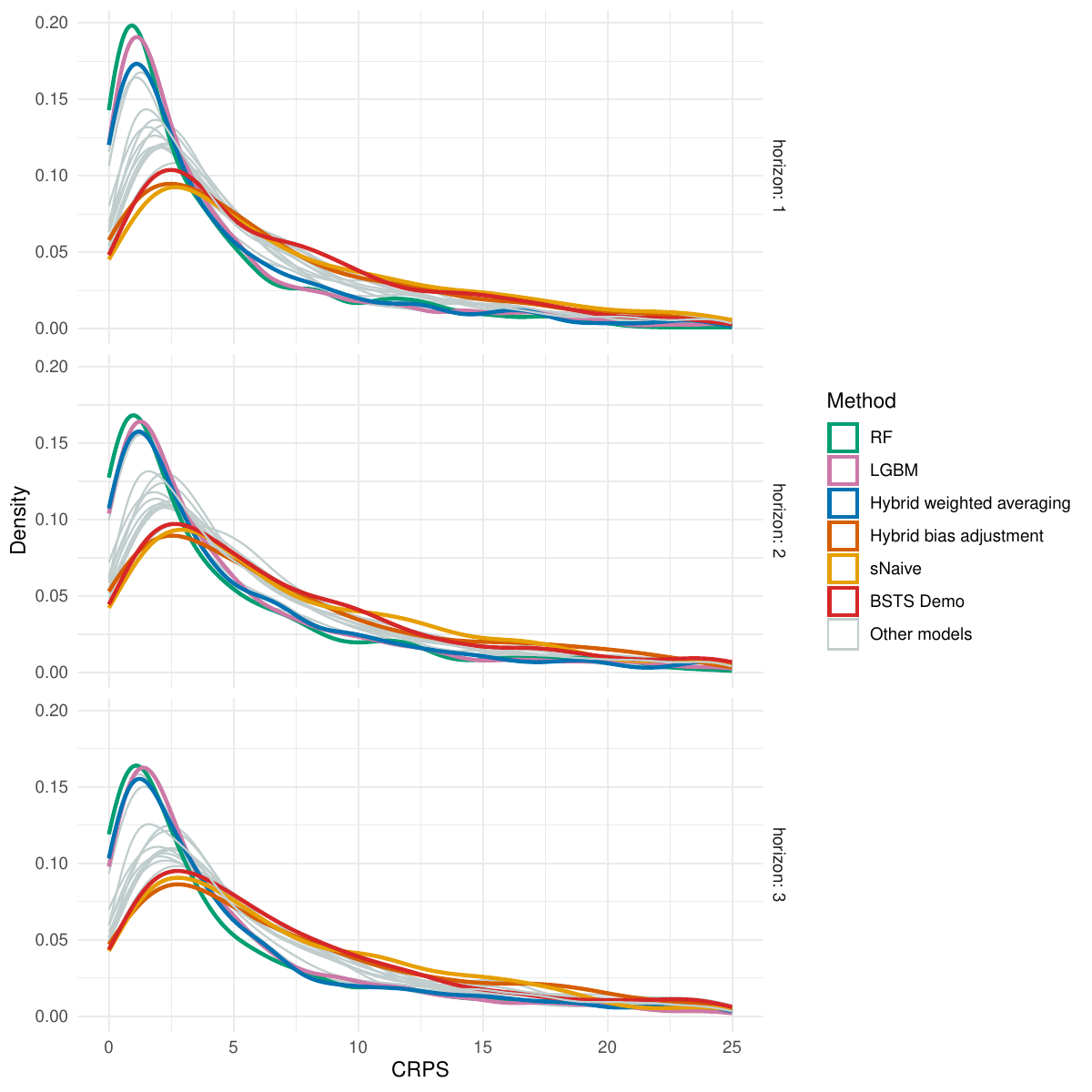}

}

\caption{\label{fig-crps_density}The distribution of CRPS values for the
top three and bottom three forecasting methods across the horizons is
presented. The methods are ranked based on their mean CRPS values, with
the top and bottom methods selected accordingly. Grey lines represent
the distribution of CRPS values for all other methods, providing a
comparative context.}

\end{figure}

Figure~\ref{fig-mase_density} and Figure~\ref{fig-crps_density}
demonstrate that both the point and probabilistic forecast accuracy
densities for the top three methods exhibit a narrower spread compared
to the bottom three methods. This indicates that the forecast errors for
these top methods are less variable and more consistently close to the
actual values across different time series than those of the bottom
three methods. The densities of all other methods, shown in grey, fall
between those of the top and bottom methods, offering broader
comparative context. Moreover, the plots show that the top methods
maintain consistent performance across forecast horizons. However, it is
noteworthy that the right tail of the density curves for RF and LGBM
becomes more volatile as forecast errors increase, particularly at
forecast horizon 3. This volatility may suggest that, while these top
two methods often deliver consistently strong performance, there remain
some uncertainties with specific time series that these methods are
unable to capture effectively. In contrast, the Hybrid Weighted
Averaging method shows a smoother tail, reflecting that it captures this
variability more effectively compared to RF and LGBM.

\subsection{Forecast performance and computational
efficiency}\label{forecast-performance-and-computational-efficiency}

We now focus on the computational efficiency of the forecasting methods.
In this study, computational efficiency is defined as the total runtime
required for one iteration on the first rolling origin. The runtime was
calculated based on this definition, and each method was retrained
during each iteration. For this analysis, we focused solely on methods
that generate both point and probabilistic forecasts from our candidate
methods.

We used two environments: an R Studio local implementation on a device
with an 11th Gen Intel(R) Core(TM) i5-1135G7 @ 2.40 GHz and 8 GB RAM, as
well as Google Colab on both CPU and T4 GPU devices. To compute the
runtime for combined statistical and ML methods, we averaged the runtime
of the respective underlying methods. For the proposed hybrid methods,
we added the runtime of the underlying methods to the time taken by the
proposed method to combine the forecasts.

Table~\ref{tbl-runtime} shows that, although the RF is the best ranked
method in Nemenyi test, it requires significantly more runtime compared
to the other forecasting methods. On the other hand, TimeGPT stands out
as the fastest method, outperforming all other methods in terms of
runtime while still providing reasonable forecast accuracy. This is a
notable exception, as it balances performance and computational
efficiency well.

However, it is important to note that ML methods were trained using a
normal CPU and one core due to technical challenges in the setup. With
access to GPU devices or CPUs with multiple cores, we could likely
improve the computational performance of these ML methods.

\begin{figure}

\centering{

\includegraphics[width=0.7\textwidth,height=\textheight]{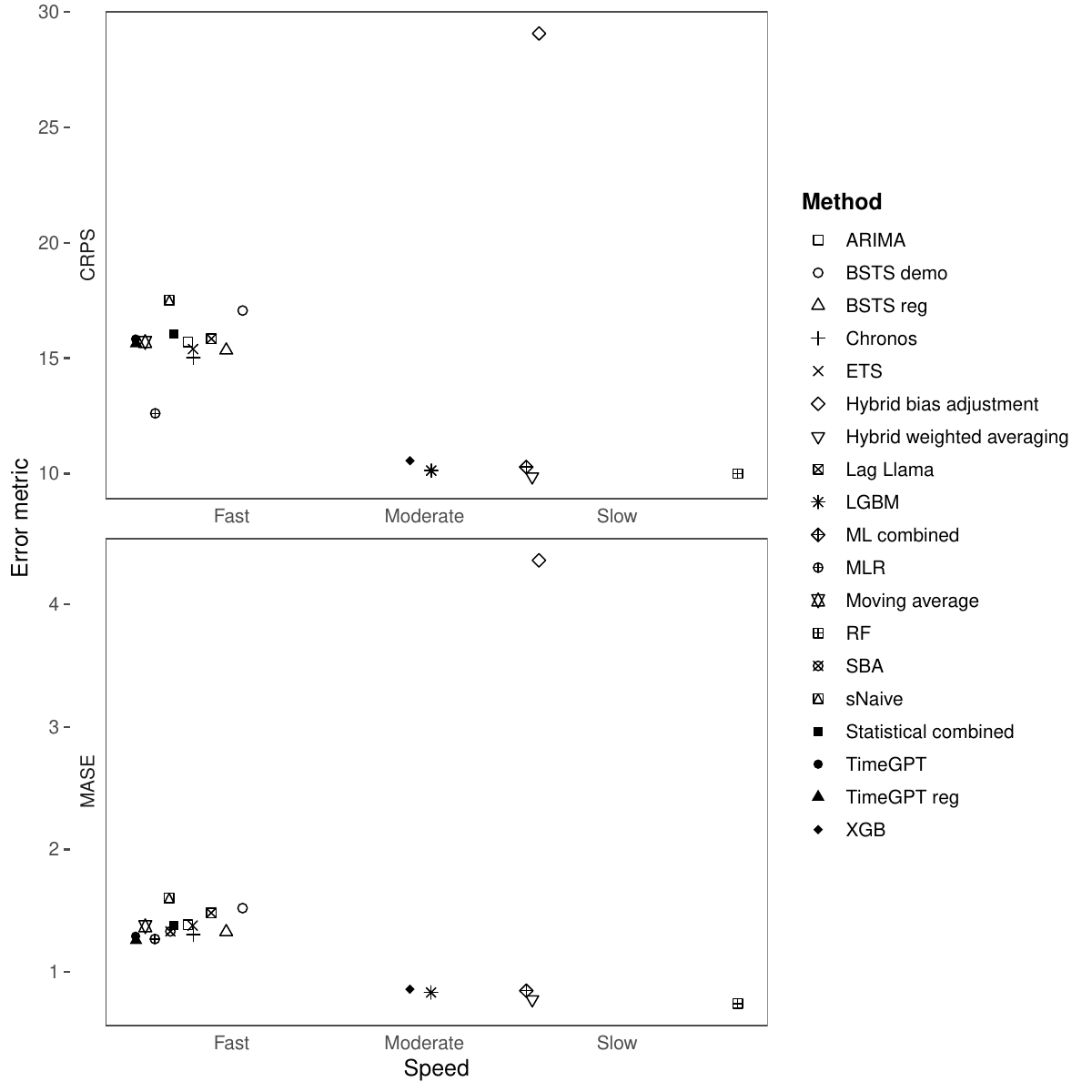}

}

\caption{\label{fig-runtime}Runtime vs.~forecast performance (The X-axis
shows the runtime speed for each method as fast, moderate, or slow).}

\end{figure}

Figure~\ref{fig-runtime} shows a clear relationship between runtime and
accuracy improvement. Most of the top-performing methods fall into the
moderate runtime category including the Hybrid Weighted Average method,
with RF (the top ranked) being the slowest method. Interestingly, the
Hybrid Bias Adjustment method also falls into the moderate runtime
category but shows a relatively high accuracy error. However, the
runtime of hybrid methods largely depends on the underlying methods
selected for combination.

From a practical perspective, choosing the right method should balance
both performance and runtime. It is a tradeoff between the extra
computational cost incurred by more sophisticated methods that can
handle uncertainties and the lower cost and simplicity of standard time
series methods.

\begin{table}

\caption{\label{tbl-runtime}Forecast performance and computational
efficiency for each forecast method are ordered based on the mean MASE.}

\centering{

\centering\begingroup\fontsize{8}{10}\selectfont

\resizebox{\ifdim\width>\linewidth\linewidth\else\width\fi}{!}{
\begin{tabular}{>{\raggedright\arraybackslash}p{14em}>{\raggedleft\arraybackslash}p{6em}>{\raggedright\arraybackslash}p{6em}>{\raggedleft\arraybackslash}p{10em}>{\raggedright\arraybackslash}p{10em}}
\toprule
Method & Mean MASE & Mean CRPS & Runtime (Minutes) & Runtime type\\
\midrule
RF & 0.743 & 9.997 & 312.50 & CPU\\
Hybrid weighted averaging & 0.775 & 9.868 & 205.90 & CPU with 4 cores\\
LGBM & 0.833 & 10.131 & 153.50 & CPU\\
ML combined & 0.847 & 10.286 & 202.83 & CPU\\
XGB & 0.859 & 10.56 & 142.50 & CPU\\
TimeGPT reg & 1.258 & 15.635 & 0.45 & Colab T4 GPU\\
MLR & 1.269 & 12.611 & 10.27 & CPU\\
TimeGPT & 1.292 & 15.831 & 0.23 & Colab T4 GPU\\
Chronos & 1.305 & 15.018 & 30.08 & Colab T4 GPU\\
BSTS reg & 1.327 & 15.342 & 47.23 & CPU with 4 cores\\
SBA & 1.331 & - & 18.23 & CPU with 4 cores\\
Moving average & 1.373 & 15.701 & 5.19 & CPU with 4 cores\\
Statistical combined & 1.378 & 16.045 & 19.99 & CPU with 4 cores\\
ETS & 1.379 & 15.397 & 29.89 & CPU with 4 cores\\
ARIMA & 1.386 & 15.703 & 27.32 & CPU with 4 cores\\
Lag Llama & 1.483 & 15.84 & 39.39 & Colab T4 GPU\\
BSTS demo & 1.521 & 17.064 & 55.73 & CPU with 4 cores\\
sNaive & 1.603 & 17.511 & 17.56 & CPU with 4 cores\\
Hybrid bias adjustment & 4.360 & 29.062 & 209.40 & CPU with 4 cores\\
\bottomrule
\end{tabular}}
\endgroup{}

}

\end{table}%

\section{Discussion}\label{sec-discussion}

\subsection{Findings}\label{findings}

Among the methods evaluated, the Hybrid Weighted Averaging method stood
out as a robust performer. In terms of mean MASE, it ranked second, and
it achieved the top rank for mean CRPS, placing it on par with the best
performing methods. Notably, the Nemenyi test revealed no significant
performance differences between the Hybrid Weighted Averaging method and
the RF method across both point and probabilistic forecasting. This
result demonstrates that the Hybrid Weighted Averaging method is a
reliable and accurate choice for forecasting contraceptive demand in
contexts where probabilistic accuracy and low variance in forecasts are
critical. Moreover, our analysis of forecast performance across multiple
time horizons found that the Hybrid Weighted Averaging method maintained
stable accuracy, which is crucial for demand planning.

One important limitation of the Hybrid Weighted Averaging method,
however, is that it becomes less suitable when the point forecast
deviates significantly from the central tendency of the probabilistic
forecast. In such cases, the Hybrid Bias Adjustment method, designed to
handle larger deviations, may be preferable. However, the bias
adjustment method produced higher errors overall. In practice, this
method can apply significant adjustments to the probabilistic forecast;
therefore, obtaining expert opinion on its estimates would be beneficial
for evaluating its performance more effectively.

The performance evaluation of both point and probabilistic forecasts
across methods showed consistent results. The MASE and CRPS analyses
reveal that top ML methods---RF, LGBM, XGB, and ML
combined---consistently outperform time series methods in both point and
probabilistic forecasting, with the RF method ranking highest on both
metrics among ML methods. This suggests that ML methods generalize well
across diverse time series patterns (smooth, erratic, lumpy,
intermittent) within the FPSC context, effectively handling the
underlying complexity of the data. Existing literature supports these
findings, indicating that ML methods are better equipped to handle
underlying uncertainties compared to time series methods
\citep{makridakis2022}. Moreover, the consistent performance of ML
methods underscores their robustness in capturing data dynamics over
time. However, the MLR method did not perform as well as other ML
methods. This discrepancy may stem from the linearity assumption
inherent in MLR, whereas real-world FPSC data likely exhibit more
complex, non-linear patterns, which MLR struggles to capture
effectively.

Despite these results, time series methods should not be entirely
discounted. For example, the SBA method, while outperformed by ML
methods, surpassed many other time series approaches in terms of MASE,
suggesting that it may be more suitable for site-level contraceptive
demand data, which often exhibit low or zero demand.
\citet{syntetos2009} also highlight the suitability of the SBA method
for such scenarios, though its limitation lies in its inability to
provide forecast distributions \citep{hyndman2021forecasting}.
Additionally, it is notable that when BSTS methods were provided with
time series based predictors and categorical and date features as
regressors, they performed significantly better than when demographic
predictors were used. A potential reason for this could be the annual
granularity of demographic predictors, whereas this study focuses on
monthly data.

Interestingly, foundational methods did not outperform top ML-based
methods. When trained as univariate methods in a zero-shot setting, they
performed similarly to time series methods, offering no clear advantage.
Although foundational methods are typically trained on large time series
datasets from various domains \citep{garza, rasul, ansari}, the time
series data observed in the FPSC context pose additional challenges such
as noise, inaccuracy, and incompleteness \citep{bearak2020}. This
highlights the need for pretraining these methods on time series data
from the humanitarian sector, which shares similar data challenges.
\citet{carriero} found that foundational methods perform better with
stationary time series, and they emphasize the importance of
incorporating external factors such as expert knowledge. Our study
corroborates this by showing that the incorporation of external
regressors significantly improved the forecasting performance of the
TimeGPT method.

Furthermore, the current methods applied to contraceptive demand
planning demonstrate that both the moving average and demographic
methods underperform, with the demographic method being the worst. One
possible reason for this is that family planning indicators are often
based on assumptions \citep{akhlaghi2013}, and these indicators are
calculated at the national or global level \citep{new2017}, making them
less reflective of local patterns. Additionally, the demographic method
provides estimates of total need rather than consumption, which can lead
to discrepancies between estimates and actual consumption
\citep{akhlaghi2013}. Additionally, our analysis of forecast performance
across multiple time horizons revealed that method performance remained
stable over time. Notably, the USAID competition submissions were not
able to outperform the top five methods in our study.

Finally, we assessed the trade-off between computational efficiency and
forecast accuracy. While RF achieved the high accuracy, it also demanded
greater computational resources. The Hybrid Weighted Averaging, LGBM,
and XGB methods offered a balanced solution, delivering high accuracy
with moderate computational demands. time series, Bayesian, and
foundational methods were computationally efficient but less accurate.
TimeGPT with external regressors, though not the most accurate, achieved
a balance between accuracy and efficiency, making it suitable for
resource-constrained contexts where moderate accuracy is acceptable.

In practice, healthcare sites generate forecasts monthly, and thus,
moderately efficient methods like LGBM are often a suitable choice.
LGBM's track record in forecasting competitions like M5
\citep{makridakis2022}, as well as its strong performance in this study,
further support its practical applicability. Accordingly, our proposed
hybrid combination approach could be employed to combine judgment
forecasts with probabilistic forecasts generated using LGBM.

\subsection{Managerial implications}\label{managerial-implications}

Demand forecasting for contraceptives in developing countries is a
critical managerial task, given the volatile and unpredictable nature of
demand. However, many field-level staff still rely on basic methods like
moving averages or demographic projections, which often fall short in
addressing these complexities. Our research underscores the need to
transition to advanced probabilistic forecasting approaches that provide
a range of potential outcomes rather than a single-point estimate. This
shift can enable field-level staff to better anticipate demand
variability and uncertainty.
\textcolor{ForestGreen}{Thus, our proposed model can support procurement strategies by providing probabilistic forecasts that help optimize order quantities based on demand uncertainty. This enables field level staff to reduce both overstocking and understocking risks by quantifying the uncertainty. Additionally, the model enhances FPSC resilience by allowing inventory levels to be adjusted dynamically based on forecast distributions, improving responsiveness to demand fluctuations while ensuring consistent contraceptive availability.}

Additionally, our findings highlight the importance of integrating
domain expertise with ML forecasts to address the limitations of purely
data-driven approaches. The variability in ML performance, particularly
in capturing extreme demand patterns, points to the value of a hybrid
approach. By allowing expert judgement to refine ML outputs, this method
improves transparency and ensures alignment with the specific goals of
FPSC. Field-level staff can therefore benefit from actionable insights
while avoiding the ``black box'' nature of many advanced forecasting
methods.

To further aid field-level staff, we developed a practical guideline
(see Table~\ref{tbl-guide} in \textbf{Appendix 2}) that compares various
forecasting methods tested in our study. This resource builds on prior
frameworks, such as the \emph{Contraceptive Forecasting Handbook} by
\citet{familyplanninglogisticsmanagement2000}, but goes further by
incorporating advanced techniques like Bayesian modeling and hybrid
methods tailored for uncertain demand environments. This guideline
serves as a roadmap for field-level staff to select the most suitable
method for their operational context, improving decision-making quality
and efficiency.

Lastly, one of the broader implications of our study is its potential
for replication in other sectors. The adaptability of our proposed
method means that it can be applied to other humanitarian or public
health contexts that deal with volatile demand patterns, such as food
aid distribution or medical supply chains. The ability to generalize
this approach across various sectors ensures that field-level staff in
different industries can also benefit from improved forecasting
practices, thereby increasing the overall reliability and resilience of
their supply chains.

\subsection{Limitations and future
directions}\label{limitations-and-future-directions}

While our study provides valuable insights into contraceptive demand
forecasting using a variety of methods, certain limitations need to be
acknowledged. First, although we compared the point and probabilistic
forecast performance across different methods ranging from statistical,
Bayesian, ML, to foundational, we did not conduct a detailed diagnostic
analysis of how each method behaves in the presence of volatile time
series, such as those typical of contraceptive demand. Volatile time
series can exhibit erratic patterns, discontinuations, and unexpected
spikes, complicating the forecasting process.

There are two main challenges in addressing this issue. First, methods
trained in a global setting (where one method handles all series) allow
for easier diagnostics. However, methods trained in a local setting
(where one method is fitted per time series) make diagnostic processes
significantly more complicated due to the large number of methods
involved. Second, there is no standard diagnostic framework that applies
across different model families, making it difficult to compare models
with varied structures. Future research should explore the development
of a standardized diagnostic framework for diverse forecasting models,
particularly in the context of contraceptive demand, as such a framework
could improve our understanding of how models behave under real-world
complexities.

Another limitation is that our linear equal-weighted forecasts did not
perform as well as expected. This may be due to the assumption that all
forecasts were well-calibrated, and thus their combination would be too.
However, the combined forecasts may have been miscalibrated, resulting
in lower performance. This issue applies to our proposed methods as
well. While some research on forecast calibration exists, such as the
work by \citet{ranjan}, further investigation is needed to improve
post-calibration processes in our hybrid methods and linear pooling
approaches. Improving calibration could enhance both accuracy and
reliability in demand forecasting.

\textcolor{ForestGreen}{On the other hand, FPSC is often subject to uncertainties arising from complex demand patterns, variable lead times, and dependence on donor support}
\citep{mukasa2017}.
\textcolor{ForestGreen}{For instance, demographic factors like the age structure of a region can influence contraceptive demand.}
\citet{haakenstad2022}
\textcolor{ForestGreen}{highlighted that young women (ages 15-25) tend to prefer short-term contraceptive methods, while older, married women are more likely to use long-term methods. However, even these behaviours are heavily influenced by social and cultural beliefs}
\citep{sedgh2016}.
\textcolor{ForestGreen}{In some regions, such as India, long-term contraceptive methods are popular among younger women}
\citep{hellwig2022}.
\textcolor{ForestGreen}{Infrastructure variability also poses challenges, particularly in rural and underserved areas, where logistical constraints can affect the distribution and accessibility of contraceptive products. Furthermore,}
\citet{karimi2021}
\textcolor{ForestGreen}{highlight that rural facilities often face difficulties such as poor road conditions, inadequate storage, and delivery delays, further complicating supply chain operations. While field-level healthcare staff rely on their contextual knowledge to adjust forecasts, much of this information remains undocumented, making it difficult to systematically incorporate into forecasting models. Identifying and defining such influencing factors and integrating them into forecasting methods remains an important research area for future exploration.}

\textcolor{ForestGreen}{Another key issue in the FPSC is the presence of censored demand due to stockouts, under-reporting, or discontinuations. In our modelling process, we did not account for these scenarios. Future research should explore how to develop forecasting methods that can handle stockout data, mitigate its impact on decision-making. Moreover, addressing the challenges of cold starts (multiple origin points) and cold ends (discontinuations) in time series forecasting is crucial, as these are prevalent in FPSC and should be considered in future methods.}

Additionally, we did not consider product switching or substitution in
response to availability or accessibility issues. Unlike other supply
chains, contraceptive product substitution is challenging because each
product has unique attributes, such as effectiveness and coverage
period. Moreover, women's preferences are influenced by health
concerns---many women are reluctant to switch products they have used
long-term due to perceived health risks \citep{sedgh2014}. Younger
women, for instance, may avoid long-term contraceptives, fearing they
could affect future fertility \citep{hellwig2022}. Investigating how to
incorporate product-switching behaviors into the forecasting process is
an important area for future research.

\textcolor{ForestGreen}{Another limitation of our study is we did not conduct an exploratory analysis of the sources of expert bias in judgemental forecasting. However, understanding when and why human forecasters introduce biases is crucial for improving hybrid forecasting accuracy. Future research could address this gap by systematically analysing the conditions under which expert bias occurs, particularly in the FPSC context. This could involve access to detailed expert forecasts, identifying systematic biases, and developing mechanisms to mitigate their impact. Such insights would further refine hybrid intelligence models by improving the integration of human intuition and algorithmic precision.}

Finally, forecast distributions are just one aspect of logistics
management in contraceptive demand forecasting. Decision-makers need to
understand how to use forecast data for FPSC operations like inventory
optimization, distribution, and procurement. As \citet{raftery2016}
suggests, forecasts may only need to provide prediction intervals or
quantiles in some cases to inform decisions. Whether this approach
applies to FPSC remains an open question. Future research should explore
how to effectively communicate probabilistic forecasts and integrate
them with inventory management, assessing the practical benefits for
FPSC decision-making and improving planning and strategy formulation.

\section{Conclusion}\label{sec-conclusion}

Effective forecasting and planning within the FPSC are essential to
ensure that contraceptives are consistently and readily available to
those who need them \citep{mukasa2017}. Accurate and reliable demand
forecasting is therefore critical within the FPSC, as it supports
informed decision-making to ensure access to safe and effective
contraceptives. This, in turn, empowers individuals and communities to
make informed reproductive health choices and helps reduce the unmet
need for contraceptives \citep{ahmed2019}.

Our study points out the need to improve contraceptive demand
forecasting by combining probabilistic forecasting methods with expert
knowledge, especially within the FPSC. Current forecasting methods often
use simple methods, like moving averages or basic demographic
approaches, which don't fully capture the complexities of contraceptive
demand. These patterns are influenced by various factors, including
stockouts, product switching, and socio-demographic variables. While
system-generated forecasts are good at showing past trends, literature
shows that expert input is vital for refining forecasts in real-world
situations with incomplete data and changing demand \citep{fildes2007}.
Therefore, we propose a new framework that enhances contraceptive demand
forecasting by merging probabilistic methods with expert insights. This
combined approach offers a promising solution for dealing with the
uncertainties and complexities of contraceptive demand in developing
countries.

Our proposed hybrid method, which combines point forecasts with
probabilistic distributions, offers a promising way to improve forecasts
by incorporating expert knowledge. The hybrid weighted averaging method
strikes a good balance between accuracy and efficiency, making it
effective for adjusting probabilistic forecasts where the algorithm has
already accounted for most uncertainties. Although the hybrid bias
adjustment method showed higher error rates, it allows for important
adjustments to probabilistic forecasts using point forecasts, especially
in situations with stockouts and incomplete data, offering greater
flexibility to integrate expert judgment.

Furthermore, we review various forecasting methods, including time
series, Bayesian, foundational time series, and machine learning
methods, along with our new hybrid methods. We provide insights into the
strengths and weaknesses of these methods, their computational
efficiency, and their most appropriate use cases. This makes our study a
useful guide for forecasting contraceptive demand.

In summary, our study addresses a key gap in probabilistic forecasting
for contraceptive demand and presents a combined approach that blends
algorithmic and human expertise. The findings from this study improve
forecasting methods within the FPSC and offer practical recommendations
for better contraceptive forecasting in developing countries.

\section*{Data availability statement}\label{reproducibility}
\addcontentsline{toc}{section}{Data availability statement}

The R and Python code, along with the data required to reproduce all
results in this paper, will be made available in a GitHub repository
upon acceptance of the paper.

\section*{Acknowledgements}\label{acknowledgements}
\addcontentsline{toc}{section}{Acknowledgements}

We would like to thank Laila Akhlaghi from John Snow, Inc.~for her
advice and feedback.

\section*{Disclosure statement}\label{disclosure-statement}
\addcontentsline{toc}{section}{Disclosure statement}

The authors report there are no competing interests to declare.

\section*{References}\label{references}
\addcontentsline{toc}{section}{References}

\renewcommand{\bibsection}{}
\bibliography{bibliography.bib}

\pagebreak

\section*{Appendix}\label{appendix}
\addcontentsline{toc}{section}{Appendix}

\subsection*{Appendix 1}\label{appendix1}
\addcontentsline{toc}{subsection}{Appendix 1}

\begin{figure}

\centering{

\includegraphics[width=0.9\textwidth,height=\textheight]{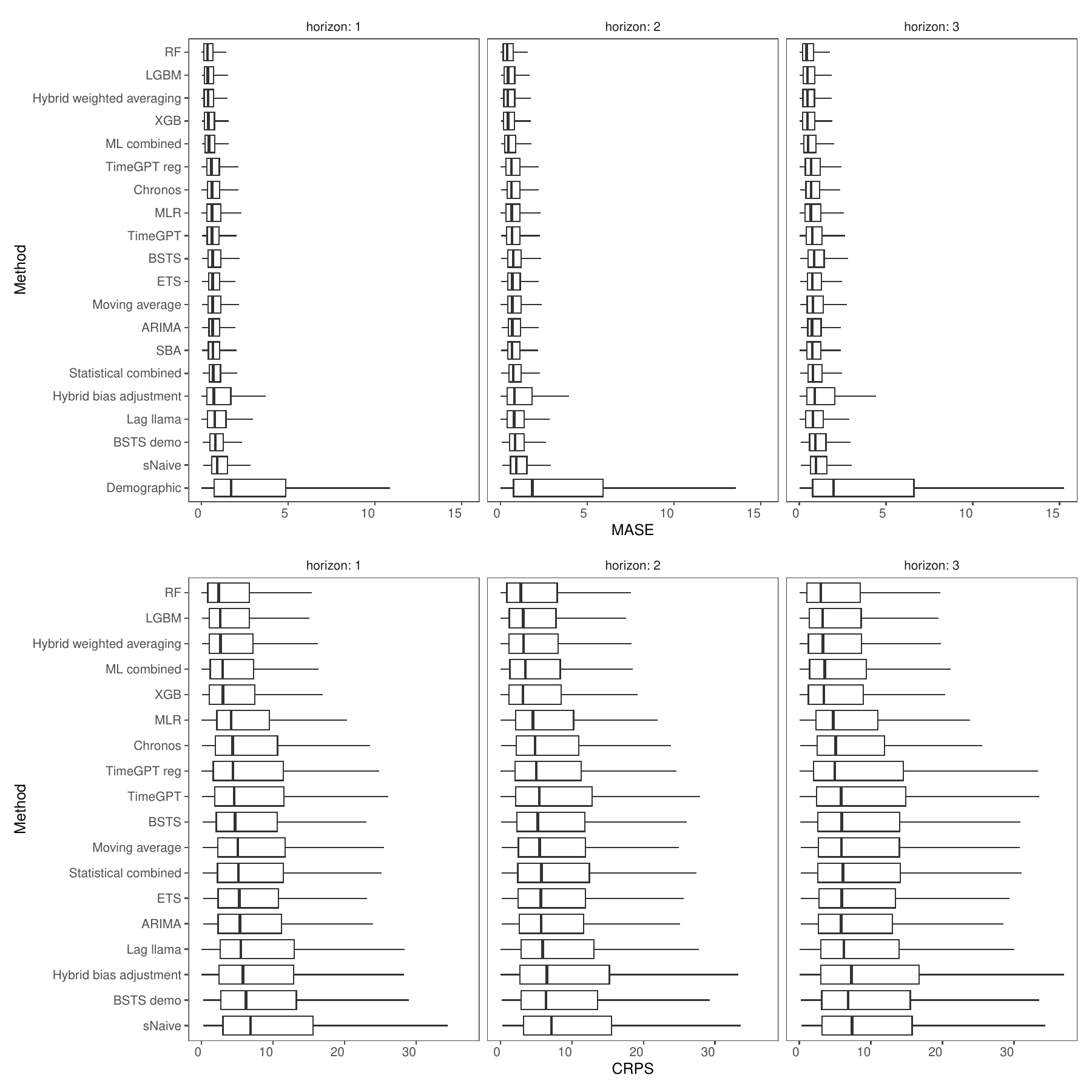}

}

\caption{\label{fig-mase_boxplot}The first panel shows the distribution
of MASE values for forecasting methods across different horizons. The
boxplots are arranged in order of the median MASE values. The second
panel shows the distribution of CRPS values for forecasting methods
across different horizons. Similarly, the boxplots are arranged in order
of the median CRPS values.}

\end{figure}%

\pagebreak

\subsection*{Appendix 2}\label{appendix2}
\addcontentsline{toc}{subsection}{Appendix 2}

\begin{landscape}\begingroup\fontsize{7}{9}\selectfont

\begin{longtable}[t]{>{\raggedright\arraybackslash}p{6em}>{\raggedright\arraybackslash}p{15em}>{\raggedright\arraybackslash}p{15em}>{\raggedright\arraybackslash}p{8em}>{\raggedright\arraybackslash}p{18em}>{\raggedright\arraybackslash}p{18em}}

\caption{\label{tbl-guide}Guidelines for method selection in
contraceptive demand forecasting.}

\tabularnewline

\toprule
Method & Strengths & Limitations & Computational efficiency & Suitable contexts & Key assumptions\\
\midrule
sNAÏVE & Simple to implement, useful as a baseline forecast model & Limited accuracy for non-stationary data, ignores trends and seasonality & Very High (minimal computational cost) & Benchmarking more advanced models, suitable for stable, short-term forecasting & Assumes future demand will be exactly the same as the last observed period\\
Moving Average & Smooths short-term fluctuations, useful for capturing general level trends & Ignores seasonality, struggles with long-term trends, lags in response to sudden changes & Very High (minimal computational cost) & Suitable for stable demand with no major seasonality; underperforms in volatile environments & Assumes future demand can be estimated by averaging past values within a chosen window\\
ETS & Captures trend, seasonality, and noise; relatively easy to interpret & Struggles with high volatility, assumes constant trend and seasonality & High (fast computational time) & Data with clear seasonality and trends, underperforms with volatile or intermittent demand data & Assumes trend and seasonality are stable over time and can be modeled separately using smoothing\\
ARIMA & Strong for univariate, stationary time series, handles seasonality well & Requires stationarity, can struggle with high volatility, requires careful model tuning & High (fast computational time) & Suitable for stationary or seasonally adjusted data, poor in volatile or intermittent demand data & Assumes data is stationary or can be made stationary through transformations (e.g., differencing)\\
SBA & Designed specifically for intermittent demand & Not suitable for continuous demand or high variability; does not handle probabilistic forecasting & Very High (minimal computational cost) & Effective for intermittent demand, with both frequent and infrequent zeroes, and with predictable inter-demand intervals & Assumes demand occurs sporadically with zero demand periods, and uses probability-based predictions for inter-demand intervals\\
\addlinespace
Multiple Linear Regression & Easy to interpret, handles multiple predictors, including external factors & Struggles with non-linearity, multicollinearity, and complex interactions & Moderate (depends on number of predictors) & Environments with clear, linear relationships between target variable and predictors & Assumes linear relationships between the dependent and independent variables\\
LightGBM & High accuracy in large, complex datasets, handles many types of predictors, efficient for large datasets & Can overfit if not carefully tuned, sensitive to noise & Moderate (more efficient than RF and XGBoost) & High-dimensional data with complex, non-linear relationships & Assumes non-linear and complex relationships that can be captured via gradient boosting algorithms\\
XGBoost & High accuracy, robust to overfitting, handles complex interactions well & Requires extensive tuning, computationally expensive compared to simpler models & Moderate (more expensive than LightGBM) & High-dimensional data, especially with complex relationships among predictors & Assumes relationships between variables can be learned through gradient boosting with proper tuning\\
Random Forest & High accuracy, handles non-linear patterns, robust for point/probabilistic forecasting & Computationally expensive, can overfit on small datasets, slow for large datasets & Low (slow, especially for large datasets) & High-dimensional datasets with complex, non-linear relationships & Assumes patterns in the data are driven by non-linear relationships learned via decision trees\\
Demographic Method & Simple, interpretable, incorporates demographic factors, useful for forecasting in new product categories or long-term planning & Poor handling of dynamic or volatile data, limited to demographic variables, not suitable for short-term forecasting & Very High (minimal computational cost) & Situations driven by demographic factors (e.g., population, age) or new product forecasting & Assumes demographic factors like population size and age are primary drivers of demand\\
\addlinespace
Bayesian Structural Time Series & Captures seasonality, trends, and structural breaks; effective for small datasets & Slow for complex models with many predictors; struggles in volatile environments & Moderate (higher with more predictors) & Data with clear seasonality, trends, or structural breaks & Assumes seasonality, trends, and causal relationships can be captured through a Bayesian framework\\
TimeGPT (with Regressors) & Highly computationally efficient, integrates external variables well & Performance degrades without external regressors, limited for non-structured data & Very High (minimal computational cost) & Low-resource environments with strong external drivers (e.g., economic factors) & Assumes external regressors are strongly correlated with demand patterns and that the time series follows stable patterns\\
Lag Llama & Captures lag effects in demand, simple to implement & Limited to contexts with strong lagged relationships; underperforms in complex scenarios & High (higher computational cost) & Situations with significant lag effects between past and future demand & Assumes demand is heavily influenced by past values with strong lag effects, and future demand can be predicted by historical lags\\
Amazon Chronos & Provides a strong baseline, simple to use & Underperforms against advanced machine learning models; limited handling of external variables & High (higher computational cost) & Univariate time series forecasting with stationary or transformed data & Assumes simple historical patterns can be extrapolated without requiring complex features or relationships\\
Hybrid Weighted Averaging Model & Combines strengths of multiple models, stable across forecast horizons & Sensitive to weight assignment, performance degrades with poor weight selection & Moderate (depends on underlying models) & Suitable for volatile or dynamic demand; can incorporate expert input for probabilistic forecasting & Assumes that multiple models capture different aspects of the demand patterns and can be effectively weighted to improve forecasting accuracy\\
\addlinespace
Hybrid Bias Adjustment Model & Corrects systematic biases in statistical models, improves forecast accuracy & Limited impact if biases are minimal, requires good bias detection & Moderate (depends on underlying models) & Ideal when systematic biases exist in forecast models; useful in dynamic or volatile demand environments & Assumes that consistent, predictable biases exist in the base models and that they can be adjusted for better forecasting\\
\bottomrule

\end{longtable}

\endgroup{}
\end{landscape}

\end{document}